\documentclass[journal]{IEEEtran}
\ifCLASSINFOpdf
\else
\fi
\usepackage{url}

\usepackage{graphicx}

\usepackage{amssymb}

\usepackage{amsthm}

\usepackage{amsmath}

\usepackage{lineno}

\usepackage{caption} 
\usepackage{url}

\usepackage{array} 

\usepackage{booktabs} 

\usepackage{bm}

\usepackage{setspace} 
\usepackage{footnote}

\usepackage{inputenc}
\usepackage{algorithm}  
\usepackage{algpseudocode}  
\usepackage{amsmath}  

\usepackage[table]{xcolor}
\usepackage{mathrsfs}
\usepackage{cite} 
\usepackage[numbers]{natbib}
\usepackage{caption}
\usepackage{bm}
\usepackage{multirow}
\usepackage{bbm}

\begin{document}
%
\title{Hybrid Multiple Attention Network for Semantic Segmentation in Aerial Images}
%
%
%


\author{Ruigang~Niu,~\IEEEmembership{Student Member,~IEEE,}
        Xian~Sun,~\IEEEmembership{Senior Member,~IEEE,}
        Yu~Tian,~\IEEEmembership{}
        Wenhui~Diao,~\IEEEmembership{}
        Kaiqiang~Chen,
        and~Kun~Fu~\IEEEmembership{}

\thanks{This work was supported by National Natural Science Foundation of China under Grants 61725105 and 41701508. \textit{(Corresponding author: Xian Sun.)}}
\thanks{R. Niu, X. Sun, Y. Tian and K. Fu are with the Aerospace Information Research Institute, Chinese Academy of Sciences, Beijing 100190, China, the Key Laboratory of Network Information System Technology (NIST), Aerospace Information Research Institute, Chinese Academy of Sciences, Beijing 100190, China, the University of Chinese Academy of Sciences and the School of Electronic, Electrical and Communication Engineering, University of Chinese Academy of Sciences, Beijing 100190, China (e-mail: niuruigang18@mails.ucas.ac.cn; sunxian@aircas.ac.cn; tianyu181@mails.ucas.ac.cn; kunfuiecas@gmail.com).}

\thanks{W. Diao and K. Chen are with the Aerospace Information Research Institute, Chinese Academy of Sciences, Beijing 100190, China and the Key Laboratory of Network Information System Technology (NIST), Aerospace Information Research Institute, Chinese Academy of Sciences, Beijing 100190, China (e-mail: diaowh@aircas.ac.cn; chenkaiqiang14@mails.ucas.ac.cn).}
}

\maketitle

\begin{abstract}
	
Semantic segmentation in very high resolution (VHR) aerial images is one of the most challenging tasks in remote sensing image understanding. Most of the current approaches are based on deep convolutional neural networks (DCNNs). However, standard convolution with local receptive fields fails in modeling global dependencies. Prior researches have indicated that attention-based methods can capture long-range dependencies and further reconstruct the feature maps for better representation. Nevertheless, limited by the mere perspective of spacial and channel attention and huge computation complexity of self-attention mechanism, it is unlikely to model the effective semantic interdependencies between each pixel-pair of remote sensing data of complex spectra. In this work, we propose a novel attention-based framework named Hybrid Multiple Attention Network (HMANet) to adaptively capture global correlations from the perspective of space, channel and category in a more effective and efficient manner. Concretely, a class augmented attention (CAA) module embedded with a class channel attention (CCA) module can be used to compute category-based correlation and recalibrate the class-level information. Additionally, we introduce a simple yet effective region shuffle attention (RSA) module to reduce feature redundant and improve the efficiency of self-attention mechanism via region-wise representations. Extensive experimental results on the ISPRS Vaihingen and Potsdam benchmark demonstrate the effectiveness and efficiency of our HMANet over other state-of-the-art methods.	

\end{abstract}


%
\IEEEpeerreviewmaketitle

\section{Introduction}
%
%
%
%

\begin{figure*}
	\centering
	\includegraphics[scale=.35]{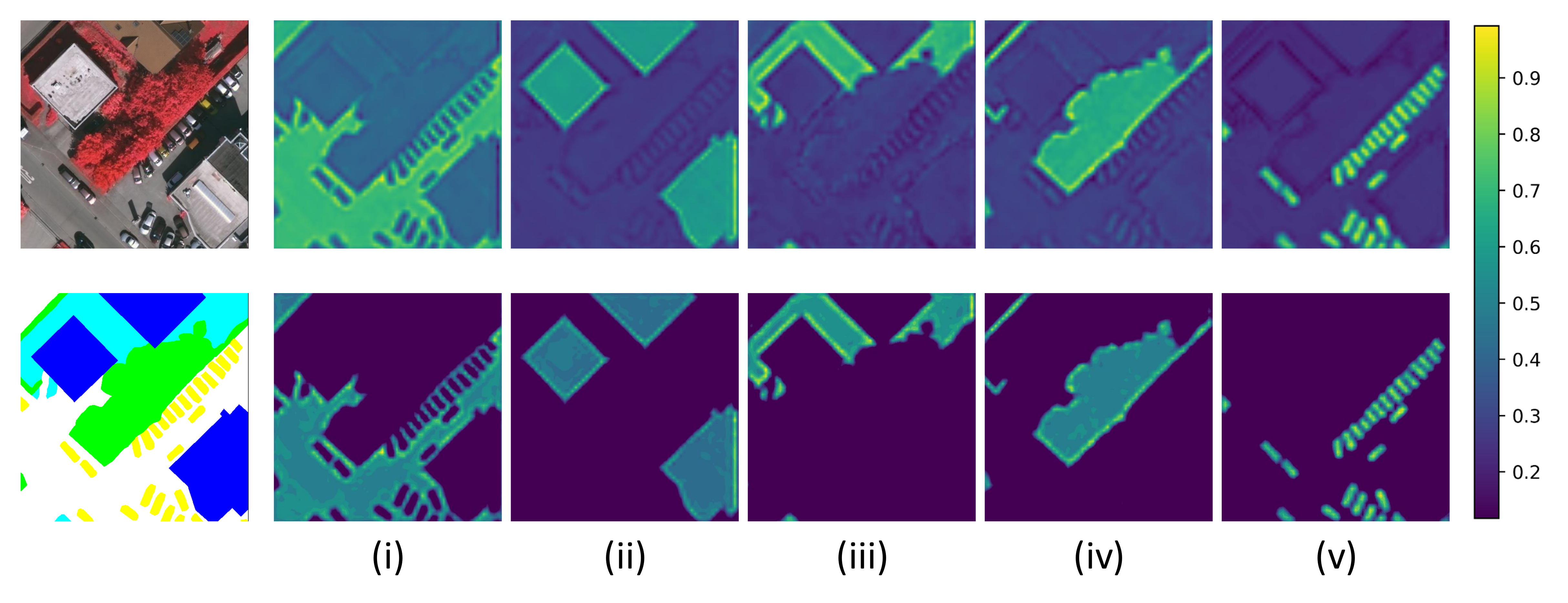}
	\caption{Visualization of Class Attention Maps corresponding to each class: (i) Imprevious surfaces. (ii) Building. (iii) Low vegetation. (iv) Tree. (v) Car. The upper row tends to represent the class attention response while the lower counterpart represents the non-negative response after non-linear activation. (Best viewed in color). }
	\label{Figure 1}
\end{figure*}

\IEEEPARstart{S}{emantic} segmentation, also known as semantic labeling, is one of the fundamental and challenging tasks in remote sensing image understanding, whose goal is to assign pixel-wise semantic class labels for a given image. In particular, semantic segmentation in very high resolution (VHR) aerial images plays an increasingly significance for its wides\-pread applications, such as road extraction \cite{maboudi2018integrating}, urban planning \cite{zhang2011mapping} and land cover classification \cite{marcos2018land}.

In recent years, Deep Convolutional Neural Networks (DCNNs) have demonstrated the powerful capacity of feature extraction and object representations compared with traditional methods in machine learning, such as Random forests (RF) \cite{pal2005random}, Support Vector Machine (SVM) \cite{gualtieri1999support} and Conditional Random Fields (CRFs) \cite{zhong2007multiple}. Particularly, state-of-the-art methods based on the Fully Convolutional Network (FCN) \cite{long2015fully} have made great progress. However, due to the fixed geometry structured, they are inherently limited by local receptive fields and short-range context information. This task is still very challenging.

To capture long-range dependencies, such as correlation coefficients between long-distance pixels, Chen \textit{et al.} \cite{chen2017rethinking} proposed atrous spatial pyramid pooling (ASPP) with multi-scale dilation rates to aggregate contextual information. Zhao \textit{et al.} \cite{zhao2017pyramid} further introduced the pyramid pooling module (PPM) to represent the feature map via multiple regions with different sizes. ScasNet \cite{liu2018semantic} aggregates multi-scale contexts in a self-cascade manner. Nevertheless, the context aggregation methods above are still unable to extract global contextual information, that is, it is unsatisfactory to cover global receptive fields by stacking and aggregating convolutional layers.

Furthermore, in order to generate dense and pixel-wise contextual information, Non-local Neural Networks \cite{wang2018non} utilizes a self-attention mechanism, which enables a single feature from any position to perceive features from all other positions. It can be seen as a matter of feature reconstruction, that is, the feature representation of each position is a weighted sum of all other counterparts. DANet \cite{fu2019dual} introduces spatial-wise and channel-wise attention modules to enrich the feature representations. Besides, several works \cite{huang2019ccnet,li2019expectation,huang2019interlaced} improve the efficiency of the self-attention mechanism to some extent. 

Semantic segmentation is essentially a pixel-by-pixel classification task, which requires the network to have a large fields-of-view. Attention-based methods have been proved to be effective ways to obtain global fields-of-view and contexts in semantic segmentation. However, the standard self-attention mechanism has many limitations in modeling effective semantic dependencies between each pixel-pair of remote sensing data of complex spectra. Hence, inspired by the success of attention-based methods above, and considering its limitations, we introduce multiple attention modules into a segmentation network to enrich the perspective of attention extraction and optimize the huge computational complexity of the self-attention mechanism.

Concretely, pixel-wise attention approaches need to generate a dense attention map to measure the relationships between each pixel-pair, which has a high computation complexity and occupies a huge number of GPU memory. Recent works \cite{chen20182,li2019expectation} have shown the fact that information redundancy is not conducive to feature representations. What's more, attention-based methods are restricted to the perspective of space and channel, ignoring category-based information, which is a key factor for semantic segmentation task. The category-based information is directly related to the last convolution of the network. In general, lack of category-based information and huge computation complexity of self-attention mechanism are two tough problems and will be elaborated below.

\begin{figure*}
	\centering
	\includegraphics[scale=0.2]{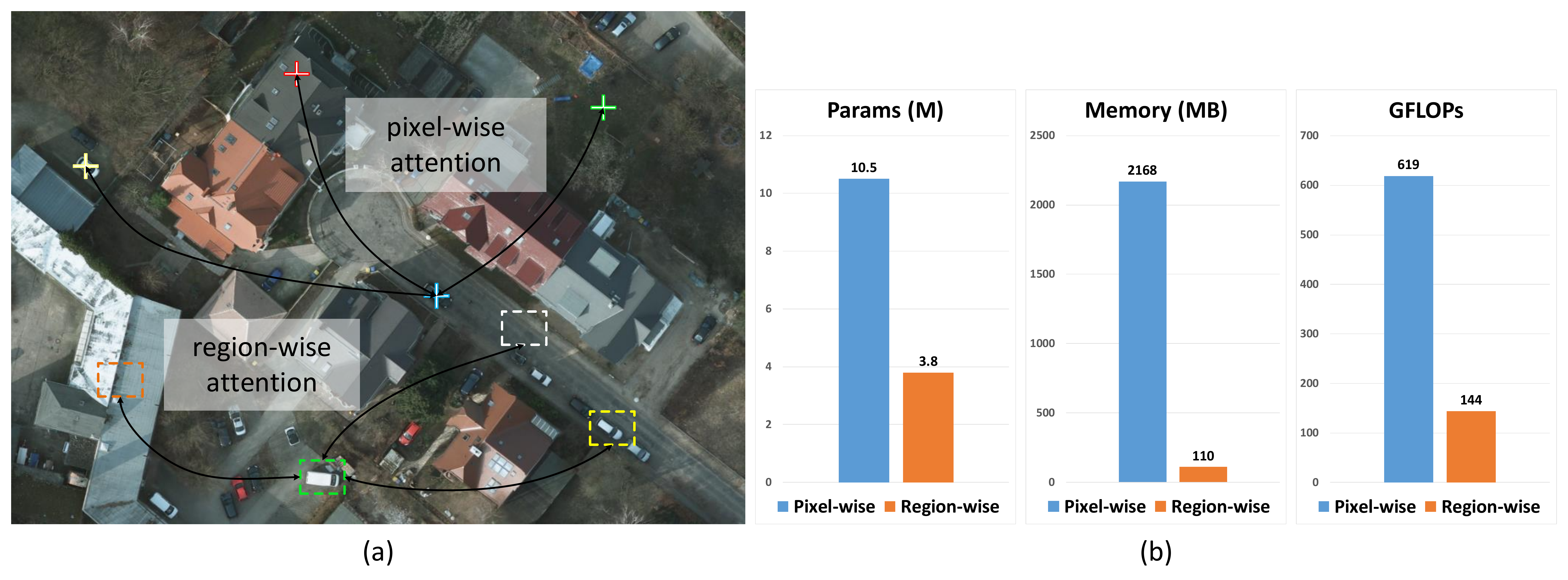}
	\caption{Intuitive understanding of pixel-wise attention and region-wise attention. The right side shows the the advantage of region-wise attention over the standard pixel-wise attention in terms of the Parameters (measured by M), GPU memory cost (measured by MB) and computation cost (measured by GFLOPs). It can be seen that the region-wise attention requires $20\times$ less GPU memory usage and reduces FLOPs by about 77\%.}
	\label{Figure 8}
\end{figure*}

On one hand, for remote sensing images of complex spectral, the class-level information is usually directly reflected between different spectra, and the input data itself has sufficient class-level information. The category-based information is embedded in different spectra, namely different channels of the input feature. But in previous works \cite{long2015fully,chen2017deeplab,chen2017rethinking,zhao2017pyramid,mi2020superpixel}, the category-based information in the general segmentation network is only reflected in the last convolutional layer, that is, the score map representing the probability that each pixel belongs to each category. In other words, through the complex feature extraction and representation of the middle stage of the network, the category-based information of the input data is already ambiguous or missing. Empirically, the lack of class-level information leads to poor object classification capabilities. Hence, different from other attention-based methods, we argue that retaining the category-based information in the middle stage of the network and extracting the corresponding attention representations. We propose a so-called category-based correlation that models class-level representation of each pixel and further calculates the relationships between categories and corresponding channels of the feature cube.  As shown in $\rm Fig.\ \ref{Figure 1}$, category-based correlation mainly focuses on exploiting contextual information from a categorical perspective, which pays more attention to the pixels of the same category during the feature reconstruction.

On the other hand, a tricky problem in remote sensing images is that the feature representations of objects with the same category are quite different in complex scenes. Therefore, the pixel-wise attention tends to extract the wrong similarity relationship between pixels, leading to serious classification errors. Besides, as illustrated in $\rm Fig.\ \ref{Figure 8}$ (b), it has a high computation complexity and occupies a huge number of GPU memory. Several works \cite{li2019expectation,chen20182} have proved that the invalid redundant information is not conducive to the feature representations. For example, as for a single feature belonging to `car' in $\rm Fig.\ \ref{Figure 8}$ (a), the pixel-wise attention method usually extracts features of all other positions, among which we actually do not need to focus on the `building' and `impervious surface', and it is more likely to extract the wrong similarity because of the complex scenes (such as in the shadow or overlapping). Aiming at the problems above, we employ a more robust region-wise attention mechanism to exploit a wider range of correlations. Empirically, region-wise representation can capture long-range contextual information between pixels in a more efficient manner.

Towards the above two issues and our corresponding solutions, we propose a novel framework, named Hybrid Multiple Attention Network (HMANet). The HMANet mainly consists of two parallel branches, one of which is the Class Augmented Attention (CAA) module embedded with the Class Channel Attention (CCA) module. Given the input feature, the CAA module first calculates category-based correlation and further generates the weighted class representation via a dense class affinity map. While the CCA module is added to adaptively recalibrate the class-level information through two linear scaling transformation functions, which efficiently helps to enhance the discriminative abilities for each class with a few parameters. The other branch of our network is the Region Shuffle Attention (RSA) module, which aims to capture region-wise global information with a shuffling operator and obtain more robust correlation between objects. Besides, compared with pixel-wise self-attention methods, the grouped region-wise representation requires $20\times$ less GPU memory usage and significantly reduces FLOPs by about 77\% with a few parameters. Finally, we concatenate the output features from each branch and the local representation, and then feed them into a convolutional layer to further generate the fine segmentation map.

\begin{figure*}
	\centering
	\includegraphics[scale=0.27]{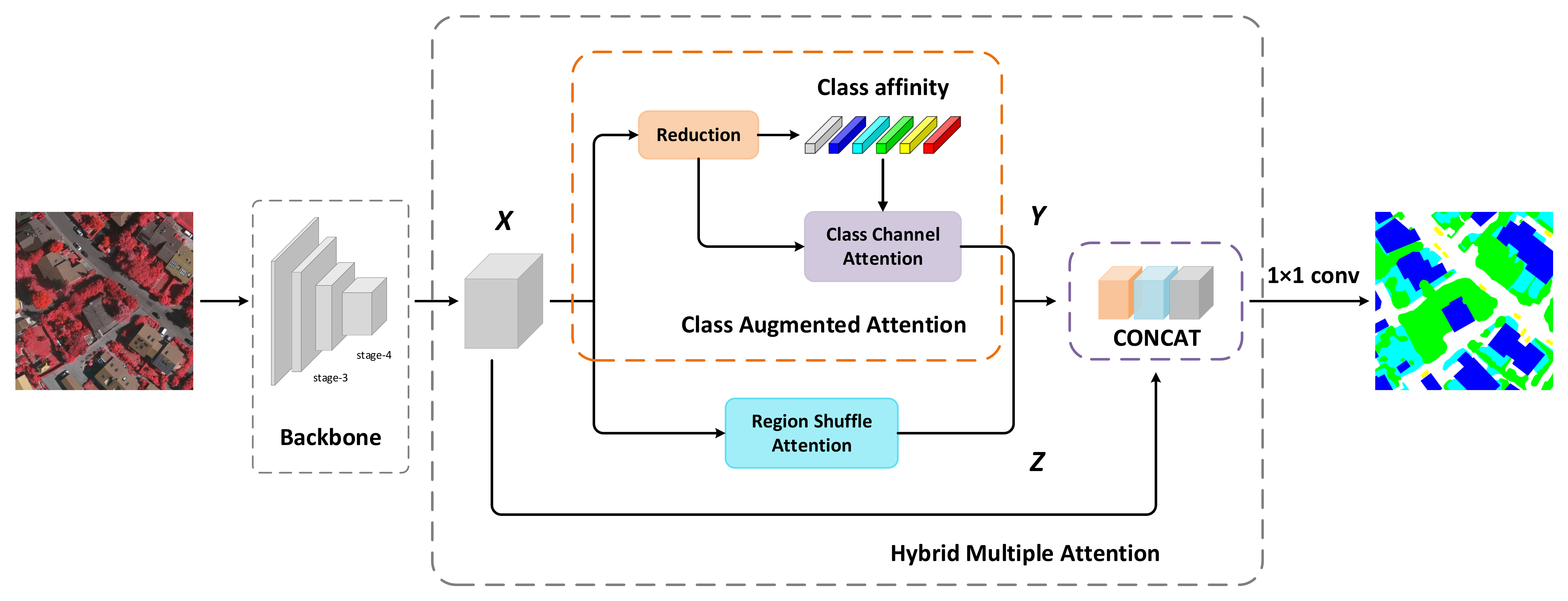}
	\caption{The pipeline of the proposed Hybrid Multiple Attention Network (HMANet). The key components are the two parallel branches, Class Augmented Attention (CAA) module embeded with Class Channel Attention (CCA) module and Region Shuffle Attention (RSA) module, which obtain the category-based correlation and region-wise contextual dependencies, respectively. Empirically, we concatenate the two output feature maps $\left\{ \rm Y, \rm Z \right\}$ and the local representation $\rm X$ to further generate the final segmentation map (Best viewed in color).}
	\label{Figure 2}
\end{figure*}

Our contributions can be summarized as follows:

\begin{enumerate}
	
	\item We present a Class Augmented Attention (CAA) module to exploit the category-based correlation between pixels and enhance the discriminant ability for each class, within which a Class Channel Attention (CCA) module is embedded to recalibrate the class-level information for better representations adaptively.
	
	\item The Region Shuffle Attention (RSA) module is proposed to capture region-wise global information and obtain more robust relationships between objects in a more efficient and effective manner.
	
	\item We propose a novel Hybrid Multiple Attention Network (HMANet) by taking advantage of the three attention modules above, which comprehensively captures feature correlations from the perspective of space, channel and category.
	
	
\end{enumerate}

The reminder of this paper is arranged as follows. Related work is briefly introduced in $\rm Section.\ \ref{section:2}$. $\rm Section.\ \ref{section:3}$ presents the details of our proposed method, including three attention modules, respectively. Experimental evaluations between our HMANet and the state-of-the-art methods, as well as ablation studies on Vaihingen dataset are provided in $\rm Section.\ \ref{section:4}$. Finally, the conclusion is outlined in $\rm Section.\ \ref{section:5}$

\section{Related Works} \label{section:2}

A full review is beyond the scope of this paper. Here, we review some recent works on semantic segmentation of nature scenes and remote sensing images. Then we turn to attention-based approach that is more relevant to our work.

\noindent\textbf{Semantic Segmentation.} Semantic segmentation is one of the fundamental tasks of image understanding. Fully Convolutional Networks (FCNs) \cite{long2015fully} based methods have made great progress in semantic segmentation by leveraging the powerful representation abilities of classification networks \cite{he2016deep,huang2017densely} pretrained on large-scale data \cite{russakovsky2015imagenet}. Several model variants are proposed to aggregate multi-scale contextual information that is vital for object perception. Concretely, D\-eepLabv2 \cite{chen2017deeplab} and DeepLabv3 \cite{chen2017rethinking} employ atrous spatial pyramid pooling (ASPP) to embed contextual representation, which consists of parallel convolutions with different dilated rates. PSPNet \cite{zhao2017pyramid} proposes a pyramid pooling module (PPM) to extract the contextual information with different scales, each of which can be considered the global representation. UNet \cite{ronneberger2015u}, RefineNet \cite{lin2017refinenet}, DFN \cite{yu2018learning}, SegNet \cite{badrinarayanan2017segnet}, DeepLabv3+ \cite{chen2018encoder}  and SPGNet \cite{cheng2019spgnet} adopt encoder-decoder structure to carefully recover the location information while retaining high-level semantic features. GCN \cite{peng2017large} utilizes global convolutional module and global pooling to harvest context information for global representations. In addition, BiSeNet \cite{yu2018bisenet} adopts efficient spatial and context path to achieve real-time semantic segmentation.

\noindent\textbf{Semantic Segmentation of Aerial Imagery.} Semantic segmentation in VHR aerial images benefits a lot from deep learning methods. For example, Mou \textit{et al.} \cite{mou2019relation} propose two network units, spatial relation module and channel relation module, to learn relationships between any two positions. TreeUNet \cite{yue2019treeunet} adopts a Tree-CNN block to transmit feature maps via concatenating connections and further fuse multi-scale representations. ScasNet \cite{liu2018semantic} proposes an end-to-end self-cascade network to improve the labeling coherence with sequential global-to-local contexts aggregation. SDNF \cite{mi2020superpixel} combines DCNNs and traditional decision forests algorithm in an end-to-end manner to achieve better classification accuracy. Marmanis \textit{et al.} \cite{marmanis2018classification} focus on semantically edge detection to restore high-frequency details and further obtain fine object boundaries. DSMFNet \cite{cao2019end} proposes a lightweight DSM fusion module to effectively aggregate depth information, within which Cao \textit{et al.} \cite{cao2019end} investigate four fusion strategies corresponding to different scenarios.

\noindent\textbf{Attention-based Methods.} Attention is widely used for various tasks, such as machine translation \cite{bahdanau2014neural,vaswani2017attention}, scene classification and semantic segmentation. Squeeze-and-Excitation Networks \cite{hu2018squeeze} recalibrated the feature representations by modeling the dependencies between channels. Non-local \cite{wang2018non} first adopts self-attention mechanism as a submodule for computer vision tasks, $i.e.$, video classification, object detection and instance segmentation. CCNet \cite{huang2019ccnet} harvests the contextual information of all the positions by stacking two serial criss-cross attention module. DANet \cite{fu2019dual} adopts similar spatial and channel attention module to generate information from all pixels, which costs even more computation and GPU memory than the Non-local operator \cite{wang2018non}. $A^2$-Nets \cite{chen20182} and Expectation-Maximization Attention Networks \cite{li2019expectation} sample sparse global descriptors to reconstruct the feature maps in an self-attention mechanism. ACFNet \cite{zhang2019acfnet} proposes a coarse-to-fine segmentation network based on attention class feature module, which can be embedded in any base network. Huang \textit{et al.} \cite{huang2019interlaced}, Yuan \textit{et al.} \cite{yuan2019object} and Zhu \textit{et al.} \cite{zhu2019asymmetric} further improve the efficiency of self-attention mechanism for semantic segmentation.

Motivated by the success of the attention-based methods above, and considering its limitations, we rethink the attention mechanism from the view of different perspectives and computation cost. Different from the previous works, we propose a Hybrid Multiple Attention to capture global contexts from the perspective of space, channel and category respectively for better feature representations. Moreover, benefiting from the multi-perspective attention mechanism and region-wise representations, HMANet is more efficient and effective than other attention-based methods. Comprehensive empirical results verify the superiority of our proposed method.

\section{Methodology}\label{section:3}
\subsection{Overview}\label{section:3.1}

\begin{figure*}
	\centering
	\includegraphics[scale=0.235]{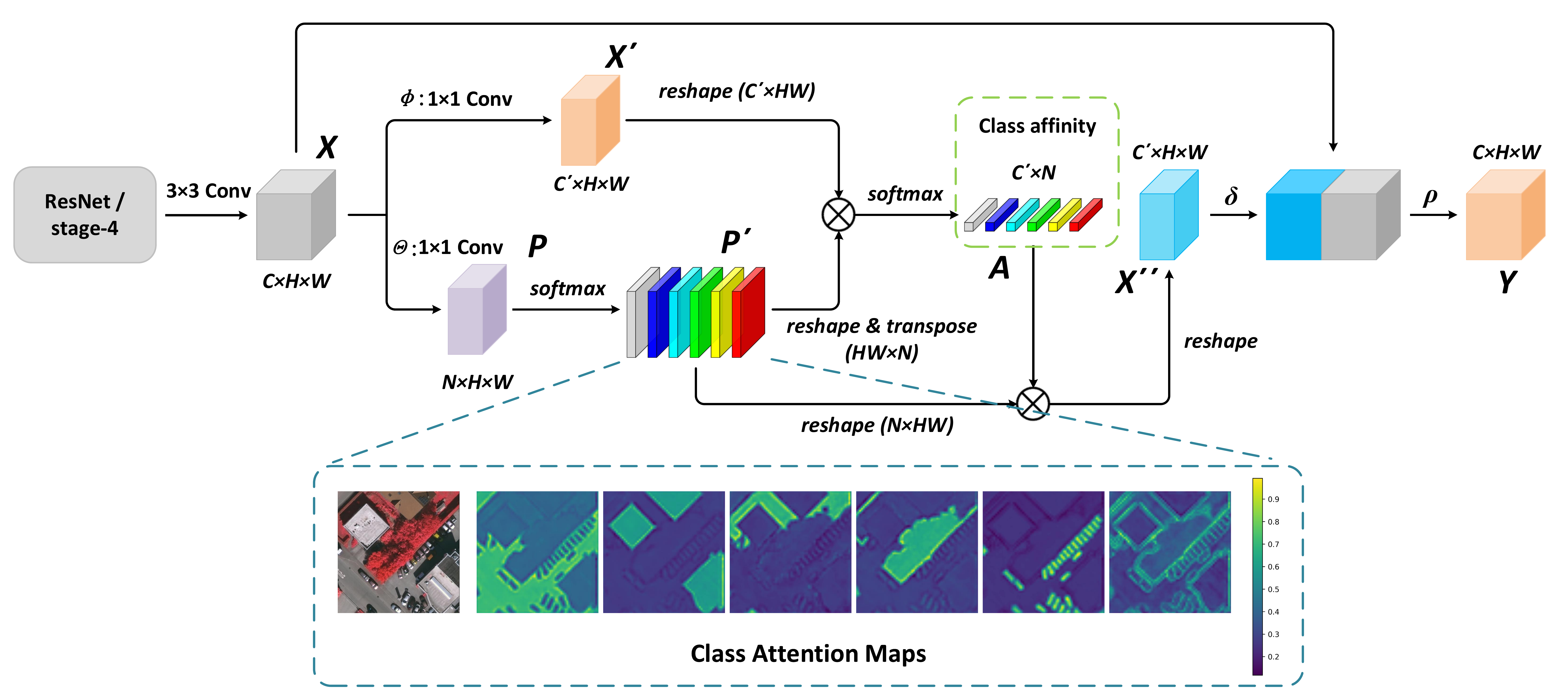}
	\caption{The details of class augmented attention module (Best viewed in color).}
	\label{Figure 3}
\end{figure*}

As shown in $\rm Fig.\ \ref{Figure 2}$, the network architecture mainly consists of three attention modules, Class Augmented Attention (CAA) module, Class Channel Attention (CCA) module and Region Shuffle Attention (RSA) module, among which CAA module and CCA module are embedded together as the upper branch of the network. The proposed CAA module aims to extract the class-level information while the CCA module improves the process of feature reconstruction via class channel weighting for better contextual representation. The lower branch of the network is the RSA module, accordingly, which greatly decreases the computational consumption and memory footprint in contrast to the original non-local block in computing long-range dependencies.

Concretely, given an input image, we first feed it into a convolutional neural network (CNN) to adaptively extract features for better representation, which is designed in a fully convolutional manner \cite{long2015fully}. We take ResNet-101 pre-trained on ImageNet dataset as our backbone. In particular, we remove the last two down-sampling operations and use dilated convolutions in stage-3 and stage-4, which is also called a multi-grid strategy for the latter, thereby retaining more spatial information and enlarging the output feature map $ \rm X$ to 1/8 of the input image without adding extra parameters. Then the features $ \rm X$ from the stage-4 of the backbone would be fed into two parallel attention branches. 

The upper branch is the CAA module embedded with the CCA module. The CAA module is designed to model the dependencies between specific categories and the corresponding features after the dimension reduction, that is, extract the similarity relationships between each category and each channel of the input feature through matrix operations. It helps to obtain a fine-grained feature representation that is more sensitive to object category information and enhance the discriminative ability of the network. The CCA module can be defined as the adaptive feature reconstruction (see $\rm Eq.\ (\ref{Equation 4})$) of class channel information, which can effectively improve the feature representation of category information. It is worth mentioning that the CCA module takes the class affinity matrix (see $\rm Eq.\ (\ref{Equation 1})$) and class attention map as the input features, both of which are generated by the CAA module, then, obtains the adaptive weighted class affinity matrix. Ideally, given the input feature map ${\rm{X}}\in{\mathbb{R}^{C\times{H}\times{W}}}$, in which $C$, $H$ and $W$ denote the number of channels, height and width of feature map respectively, the CAA embedded with CCA module can effectively extract the class-channel correlation and adaptively aggregate long-range contextual information from a category view, eventually, outputs the same size feature map $ {\rm{Y}}\in{\mathbb{R}^{C\times{H}\times{W}}}$ following the self-attention scheme \cite{wang2018non}. 

The lower branch of the network, RSA module, is proposed with the intuition of decomposing the dense point-wise affinity matrix into two sparse region-based counterparts, either of which could efficiently capture the global context in a sparser way via adaptive average pooling method. With the combination of the two affinity matrices, the RSA module could capture abundant spatial contextual information of the local input feature X then output feature ${\rm{Z}}\in{\mathbb{R}^{C\times{H}\times{W}}}$. Finally, we concatenate the output features of the two branches $ \left\{ \rm Y, \rm Z \right\}$ and the local feature representation $  \rm X$ to obtain better feature representations, then, the fused features are fed into a convolutional layer to generate the fine segmentation map.

\subsection{Class Augmented Attention} \label{section:3.2}

The self-attention mechanism is essentially a kind of matrix multiplication operation in mathematics, in which the two dimensions are the number of channels $ \left\{ C \right\} $ and the product of height and width $\left\{ H \times W \right\}$ of the input feature map respectively. The standard channel affinity matrix of size $C \times C$ can be obtained by the matrix multiplication of two inputs with dimension $C\times HW$ and $HW\times C$, such as channel attention module in DANet \cite{fu2019dual}. Intuitively, the definition of non-local operation constrains the scaling of the channel in such kind of channel attention module, that is, the query, key and value functions are eliminated during the operation. Nevertheless, it leads into category information when one of the channels $C$ is replaced by the channel corresponding to the segmentation map supervised by the ground truth, retaining the query, key and value transformation functions in the meantime.

The intuition of the proposed class augmented attention is to capture long-range contextual information from the perspective of category information, that is, to explicitly model the relationships between each category in the dataset and each channel of the input feature cube. Next, we will elaborate the process to capture class-level contextual information.

As shown in $\rm Fig.\ \ref{Figure 3}$, given a local feature ${\rm{X}}\in{\mathbb{R}^{C\times{H}\times{W}}}$, output from the $3\times 3 \rm\  conv$ after stage-4 of ResNet in our implementation, the class augmented attention module first applies two convolutional layers to generate two feature maps ${\rm{X'}}\in{\mathbb{R}^{C'\times{H}\times{W}}}$, and ${\rm{P}}\in{\mathbb{R}^{N\times{H}\times{W}}}$, respectively, where $C'$ is the reduced channel number of the local feature for less computational cost and P is the class attention map supervised by the ground-truth segmentation. For each channel $k$ in ${\rm{P}}\in{\mathbb{R}^{N\times{H}\times{W}}}$, ${\rm{P_k}}\in{\mathbb{R}^{H\times{W}}}$ is available to represent the confidence that pixels of all position $i$ belongs to class $k$, where $N$ is the number of categories. $\rm X'_u$ represents the $u$th element of $\rm X'$ along channel dimension. Then, we can further generate the class affinity map ${\rm{A}}\in{\mathbb{R}^{C'\times{N}}}$ via aggregating all the position $i$ in spatial dimension of $\rm X'$ and $\rm P$ after a softmax layer. The class affinity operation is defined as follows:

\begin{equation}
\label{Equation 1}
s_{u,k}=\sum_{i}x'_{u,i}\ \frac{e^{p_{k,i}}}{\sum_{j=1}^{N}e^{p_{j,i}}}
\end{equation}

\noindent where $s_{u,k}\in{\rm S}$ denotes the explicit class correlation between feature $\rm X'_u$ and $\rm P_k$, $u=[1,2,...,C']$, $k=[1,2,...,N]$, ${\rm S}\in\mathbb{R}^{C'\times{N}}$. Then, we apply a softmax operation along the class dimension to generate the class affinity map $\rm A$.

The final class augmented object representation can be formulated as below:

\begin{equation}
\label{Equation 2}
{\rm Y_{u}}=
\rho(\delta\sum^N_{k=1}({a_{u,k}}\cdot\frac{e^{P_{k}}}{\sum_{j=1}^{N}e^{P_{j}}})+{\rm X_u})
\end{equation}

\noindent in which $\rm Y_u$ denotes the $u$th feature plane of the output feature map ${\rm{Y}}\in{\mathbb{R}^{C\times{H}\times{W}}}$. $a_{u,k}$ is a scalar value of $s_{u,k}$ after softmax layer. Here, $\delta(\cdot)$ and $\rho(\cdot)$ are both transformation functions implemented by $1\times 1\ \rm conv \rightarrow BN \rightarrow ReLU$. The original local feature $\rm X$ is added to enhance the feature representation. The $\rm Eq.\ (\ref{Equation 2})$ indicates that the final representation of each channel is a category-based weighted sum of all channels in class attention map, which models the category-based semantic dependencies between feature maps. That is to say, the proposed CAA module improves the perception and discriminability of class-level information in a straightforward manner.

\subsection{Class Channel Attention}\label{section:3.3}

\begin{figure}
	\centering
	\includegraphics[scale=0.2]{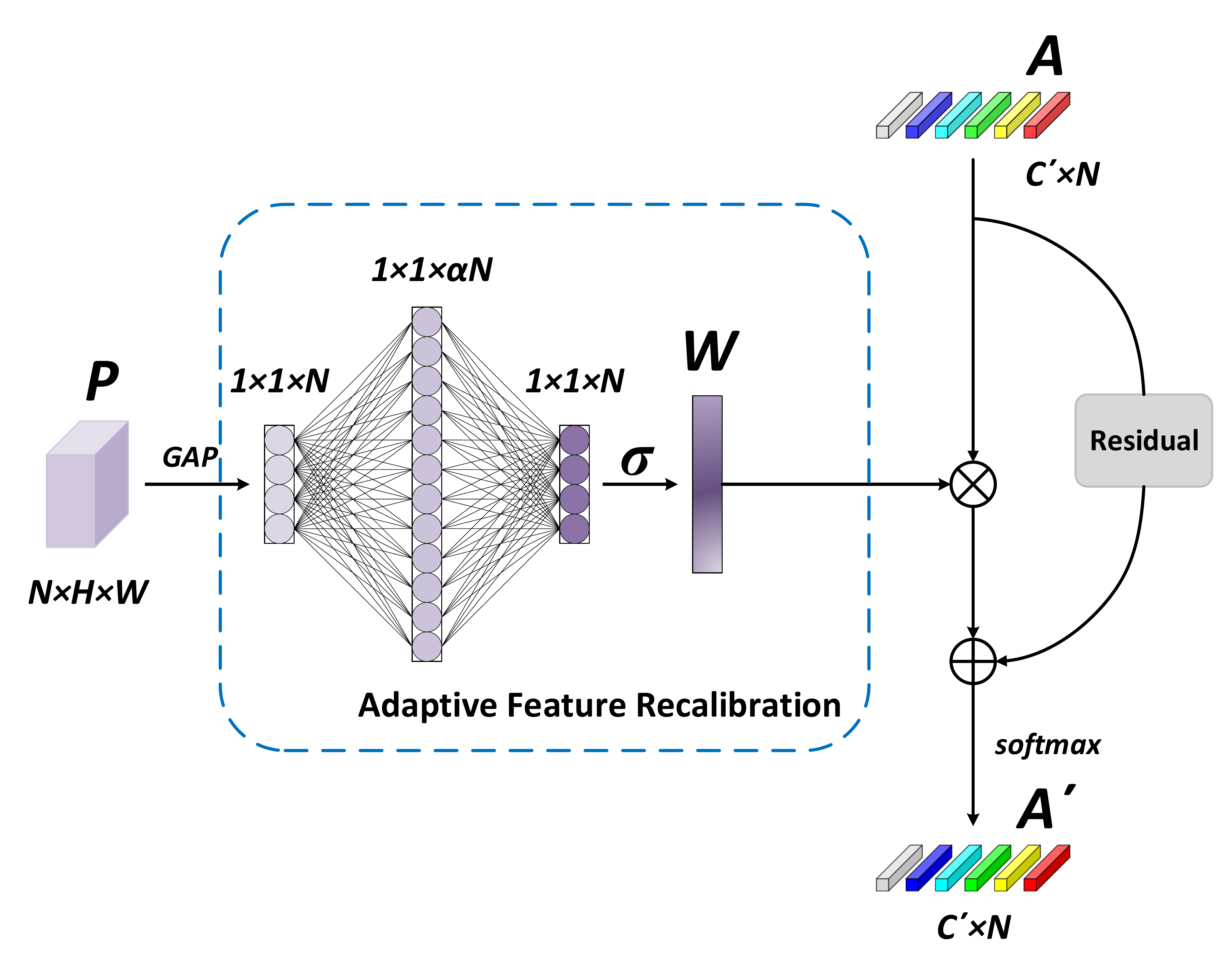}
	\caption{Diagram of class channel attention module. }
	\label{Figure 4}
\end{figure}

The high-level semantics of CNNs are empirically considered to be embedded in the channel dimension, among which each channel map of deep features can be regarded as a class-related response. Additionally, recent works \cite{hu2018squeeze, wang2019eca} have demonstrated the effectiveness of modeling channel correlation in classification and segmentation tasks. Therefore, we propose a class channel attention (CCA) module to exploit class channel dependencies and generate a new class affinity map with rich and adaptive contextual information, which is effectively embedded in the CAA module with a few parameters.

The main structure of class channel attention module is illustrated in $\rm Fig.\ \ref{Figure 4}$. Given the class attention map ${\rm{P}}\in{\mathbb{R}^{N\times{H}\times{W}}}$ and class affinity map ${\rm{A}}\in{\mathbb{R}^{C'\times{N}}}$ output from the CAA module above, the adaptive class channel statistical representations can be formulated as follows:

\begin{equation}
\label{Equation 3}
{\rm W_k}= \sigma (f_{\left\{ \rm W_1,\rm W_2 \right\}} \cdot ({ GAP} (\frac{e^{P_k}}{\sum_{j=1}^{N} e^{P_j}})))
\end{equation}

\noindent where $ GAP(P_k)=\frac 1 {HW} \sum_{i=1}^H \sum_{j=1}^W P_k(i,j) $ is the channel-wise global average pooling (GAP) to generate class-related statistics and $\sigma$ is the Sigmoid activation. Let $x = GAP(P_k)$, the key adaptive feature recalibration function is defined as:

\begin{equation}
\label{Equation 4}
f_{\left\{\rm W_1,W_2 \right\}}(x)={\rm W_2\eta (W_1}x)
\end{equation}

\noindent in which ${\rm W_2} \in \mathbb{R}^ {N\times \alpha N}$ and ${\rm W_1} \in \mathbb{R}^ {\alpha N\times N}$ and $\eta$ denotes the ReLU function. Concretely, ${\rm W_1(\cdot)}$ and ${\rm W_2(\cdot)}$ are two linear fully connected transformations, $i.e.$, dimensionality adjustment layers with ratio $\alpha$ (this parameter value will be discussed in $\rm Section \ \ref{section:4.4.2}$) to augment and squeeze the representations of category information in the channel dimension, respectively. Noted that we opt to employ a simple ReLU function to ensure the non-linearity of the model and limit the complexity following the Squeeze-and-Excitation Networks \cite{hu2018squeeze}.

The final output of the CCA module is obtained by recalibrating $ \rm A$ with the weighted factor $\rm W_k$ and the original class affinity map:

\begin{equation}
\label{Equation 5}
{\rm A'}= softmax(\sum_{i=1}^N (\gamma \rm W_i+1) A_i)
\end{equation}

\noindent where $\gamma$ is a learnable parameter initialized to $0$. The residual connection is added to retain the original representation (see``+1'' in $\rm Eq.\ \ref{Equation 5}$), thus, it can be integrated into the standard CAA module above without breaking its initial behavior, which efficiently helps to enhance the feature adaptive recalibration of class information.

\subsection{Region Shuffle Attention}\label{section:3.4}

\begin{figure}
	\centering
	\includegraphics[scale=0.19]{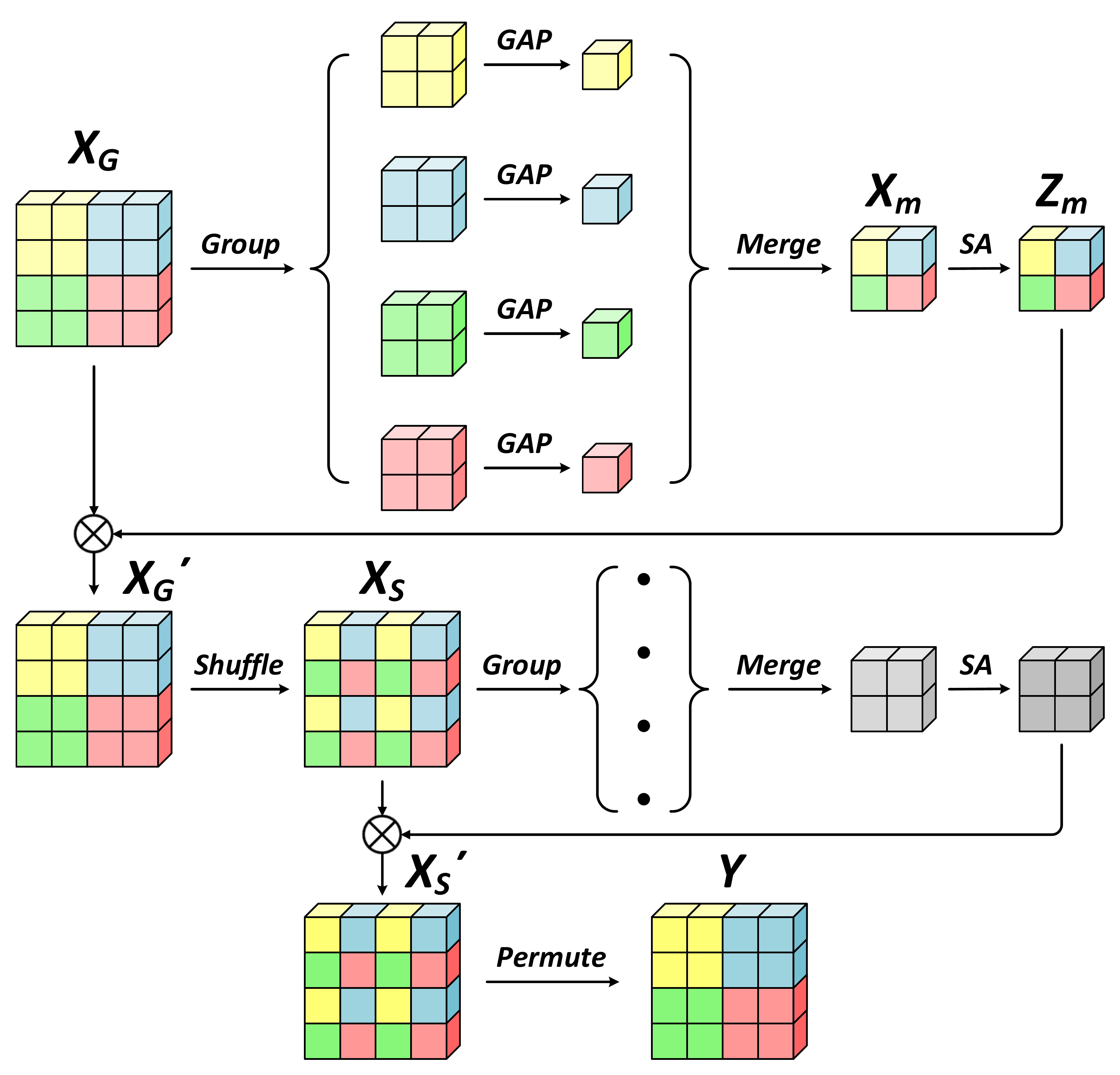}
	\caption{An example of region shuffle attention when the numbers of partitions $G$ and positions in each patition $P$ are both $2$. }
	\label{Figure 5}
\end{figure}

Attention-based neural networks, in terms of spatial point-wise correlation representations, mainly aim to capture long-range contextual dependencies through a self-attention mechanism or its variants, eventually, generating a dense affinity matrix. Even for smaller feature maps, the point-wise affinity matrix obtained will take up a lot of (GPU) memory. Hence, the key point of the proposed region shuffle attention is to harvest the region-wise dependencies as well as its counterparts after recombination in a sparse and efficient manner. We illustrate our approach via a simple schematic in $\rm Fig.\ \ref{Figure 5}$.

\noindent\textbf{Region Representations.} We partition the input feature maps into regions via a permutation operation, each of which is fed into an adaptive global average pooling layer to obtain the region representations afterwards. Then, we merge the point-wise representations of the regions to generate a sparse representation of the whole input feature. Therefore, the self-attention on the original input features can be effectively replaced with the same attention towards the merged counterparts for convenience.

\noindent\textbf{Shuffle Attention Representations.} Despite the self-attention on the merged feature that can empirically capture long-range contextual information from all positions, the pixel-to-pixel connections are still ambiguous. In order to exploit more explicit contextual dependencies from a regional perspective, we apply a shuffle attention to alternately pool the corresponding sub-regions and compute its self-attention representations, respectively, achieving a complementary representation of spatial information. Further experiments show that the cascade of attention weighted representations of the two sub-regions can effectively enhance the contextual dependencies, superior to the pixel-wise non-local operator.

As illustrated in $\rm Fig.\ \ref{Figure 5}$, we first divide the input feature $\rm X $ into $G$ partitions and each partition contains $P$ positions, where each ${\rm X_{G, p}} \in \mathbb{R}^ {C\times P}$ is a subset of $\rm X_G $. Then, we merge the point statistics after global average pooling to obtain the sparse representation ${\rm X_m} \in \mathbb{R}^ {C\times G}$. We apply self-attention on $\rm X_m$ following the non-local operation \cite{wang2018non} as below:

\begin{equation}
\label{Equation 6}
{\rm A_m} = softmax(\frac{\theta(\rm X_m)^T \phi(\rm X_m)}{\sqrt d})
\end{equation}
\begin{equation}
\label{Equation 7}
{\rm Z_m} = w {\rm A_m}g(\rm X_m)+X_m
\end{equation}

\noindent where $ {\rm A_m} \in \mathbb{R}^ {G\times G}$ is a sparse affinity matrix based on global information and ${\rm Z_m} \in \mathbb{R}^ {C\times G}$ is the weighted output features. Here, $\theta(\cdot)$ and $\phi(\cdot)$ are both transformation functions implemented by $1\times 1\ \rm conv \rightarrow BN \rightarrow ReLU$ while $g(\cdot)$ represents $1\times 1\ \rm conv$. $w$ is a learnable parameter initialized to $0$.

The regional weighted representation $\rm X_G'$ can be obtained by region-wise multiplication of $\rm Z_m$ and $\rm X_G$. We apply another permutation to regroup the representations, then, the feature $ \rm X_S$ would be fed into the same region-wise attention block to generate the final representations $\rm Y$.

Compared with standard self-attention mechanism, our approach greatly  reduces the complexity in time and space from $\mathcal{O}((H\times W)^2C)$ to $\mathcal{O}(2(\frac{1}{G_h^2G_w^2}+\frac{1}{P_h^2P_w^2})(H\times W)^2C)$, where $G_h$ and $G_w$ are the number of partitions along height and width dimensions while each partition contains $P_h$ and $P_w$ pixels, respectively.

In general, the proposed region shuffle attention module makes up for the deficiency of non-local block that it is a huge consumption of memory footprint. Additionally, it can be plugged into any existing architectures at any stage without breaking its initial performance, and optimized in an end-to-end manner.

\subsection{Hybrid Multiple Attention Network}\label{section:3.5}

\noindent\textbf{Integration of Attention Module.} In order to take full advantage of three proposed attention modules, we further aggregate the CAA module embedded with the CCA module (the upper branch illustrated in $\rm Fig.\ \ref{Figure 2}$) and the RSA module (the lower branch) in an cascading and parallel manner, both of which is concatenated with the local feature. Eventually, the feature after concatenation would be fed into the last convolution to generate the final segmentation map.

\noindent\textbf{Loss Function.} Besides the conventional multi-class cross entropy loss $\mathcal L_{ce}$, we use the auxiliary supervision $\mathcal L_{aux}$ after stage-3 to improve the performance and make it easier to optimize following PSPNet \cite{zhao2017pyramid}. The auxiliary loss can be formulated as:
\begin{equation}
\label{Equation 12}
\mathcal L_{aux}= 
-\frac{1}{BN}\sum_{i=1}^{B}\sum_{j=1}^{N}\sum_{k=1}^{K}\mathbb{I}(g_j^i=k)\log (\frac{\exp(p_{j,k}^i)}{\sum_{m=1}^{K} \exp(p_{j,m}^i)}) 
\end{equation}

\begin{equation}
\label{Equation 13}
\mathbb{I}(g_j^i=k)=
\begin{cases}
1, & g_j^i=k,\\
0, & otherwise 
\end{cases}
\end{equation}

\noindent where $B$ is the mini batch size; $N$ is the number of pixels in each batch; $K$ is the number of categories; $p_{j,k}^i$ is the prediction after ResNet-stage-3 of the $j$-th pixel in the $i$-th patch for the $k$-th class; $\mathbb{I}(g_j^i=k)$ is an indicator function as illustrated in $\rm Eq.\ \ref{Equation 13}$, it takes 1 when the ground truth of the $j$-th position in the $i$-th patch ($i.e.$ $g_j^i$) belongs to the $k$-th class, and 0 in other cases.

The class attention loss $\mathcal L_{cls}$ from CAA module is also employed as an extra auxiliary supervision. Likewise, the class attention loss can be formulated as:
\begin{equation}
\label{Equation 14}
\mathcal L_{cls}= 
-\frac{1}{BN}\sum_{i=1}^{B}\sum_{j=1}^{N}\sum_{k=1}^{K}\mathbb{I}(g_j^i=k)\log (\frac{\exp(a_{j,k}^i)}{\sum_{m=1}^{K} \exp(a_{j,m}^i)}) 
\end{equation}
\noindent where $a_{j,k}^i$ is the response value of the class attention map of the $j$-th pixel in the $i$-th patch for the $k$-th class; other definitions are the same as above.

Finally, we use three parameters to balance these loss as follows:

\begin{equation}
\label{Equation 8}
\mathcal L= 
\lambda_{1} \mathcal L_{ce} + \lambda_2 \mathcal L_{cls} + \lambda_3 \mathcal L_{aux}
\end{equation}

\noindent where $\lambda_1$, $\lambda_2$ and $\lambda_3$ are set as $1$, $0.5$ and $0.4$ to balance the loss. Noted that the ablation studies for the three loss functions and the sensitivity of the model to the choice of the weight values will be elaborated in $\rm Section \ \ref{section:4.4.7}$.

\section{Experiments}\label{section:4}
To validate the effectiveness of our proposed method, we conduct extensive experiments on two state-of-the-art aerial image semantic segmentation benchmarks, $i.e.$, ISPRS 2D Semantic Labeling Challenging for Vaihingen \cite{vaihingen} and Potsdam \cite{potsdam}, consisting of very high resolution true ortho photo (TOP) tiles and corresponding digital surface models (DSMs) derived from dense image matching techniques. In this section, we first introduce the datasets and implementation details, then we perform extensive ablation experiments on the ISPRS Vaihingen dataset. Finally, we report our results on the two datasets.

\subsection{Datatsets}\label{section:4.1}
\noindent\textbf{Vaihingen.} The Vaihingen dataset contains 33 orthorectified image tiles (TOP) mosaic with three spectral bands (red, green, near-infrared), plus a normalized digital surface model (DSM) of the same resolution. The dataset has a ground sampling distance (GSD) of 9 cm, with an average size of $2494\times 2064$ pixels, which involves five foreground object classes and one background class. We use the benchmark organizer defined 16 images for training and 17 to test our model following the previous works \cite{mou2019relation, maggiori2017high, volpi2016dense, sherrah2016fully, marcos2018land}. Noted that we do not use DSM in our experiments.

\noindent\textbf{Potsdam.} The Potsdam 2D semantic labeling dataset is composed of 38 high resolution images of size $6000\times 6000$ pixels, with a ground sampling distance (GSD) of 5 cm. The dataset offers NIR-R-G-B channels together with DSM and normalized DSM. There are 24 images in training set and 16 images in test set, which have 6 foreground classes corresponding to the Vaihingen benchmark.

\subsection{Evaluation Metrics}\label{section:4.2}
To evaluate the performance of the proposed network, we calculate the $F_1$ score for the foreground object classes with the following formula:

\begin{equation}
\label{Equation 9}
F_1 = (1\times \beta^2) \cdot \frac{precision \cdot recall}{\beta^2 \cdot precision + recall}
\end{equation}

\noindent where $\beta$ is the equivalent factor between precision and recall and is set as $1$. Intersection over union (IoU) and overall accuracy (OA) are defined as:

\begin{equation}
\label{Equation 10}
{\rm{IoU}} = \frac{TP}{TP+FP+FN}
\end{equation}

\begin{equation}
\label{Equation 11}
{\rm{OA}} = \frac{TP+TN}{N}
\end{equation}

\noindent in which $TP$, $TN$, $FP$ and $FN$ are the number of true positives, true negatives, false positives and false negatives, respectively. $N$ is the total number of pixels.

Notably, overall accuracy is computed for all categories including background for a comprehensive comparison with different models. Additionally, the evaluation is carried out using ground truth with eroded boundaries provided in the datasets following previous studies.

\subsection{Implementation Details}\label{section:4.3}
We use ResNet-101 \cite{he2016deep} pretrained on ImageNet \cite{russakovsky2015imagenet} as our backbone and employ a poly learning rate policy where the initial learning rate is multiplied by $1-(\frac{iter}{max\_ iter})^{power}$ with $power = 0.9$ after each iteration following the prior works \cite{chen2017deeplab, li2019expectation, huang2019ccnet}. And we utilize stochastic gradient descent (SGD) optimizer with the initial learning rate 0.01 for training. Momentum and weight decay coefficients are set to 0.9 and 0.0005 respectively. We replace the standard BatchNorm with InPlace-ABNSync \cite{rota2018place} to synchronize the mean and standard-deviation of BatchNorm across multiple GPUs. For the data augmentation, we apply random horizontal flipping, random scaling (from 0.5 to 2.0) and random crop over all the training images. The input size for all datasets is set to $512\times 512$. We employ $4\times$ NVIDIA Tesla K80 GPU for $80k$ iterations and batch size is $4$. For semantic segmentation, we choose FCN (VGG-16) \cite{long2015fully} pretrained on ImageNet as our baseline, and we also utilize ResNet-101 \cite{he2016deep} baseline for further comparison experiments.

\begin{table}[t]
	\centering	
	\caption{Comparisons of different weight parameters $\lambda_2$.}
	\label{Table 10}
	\begin{tabular}{ccccccc}
		\toprule
		\textbf{$\lambda_2$} & 0.2 & 0.3 & 0.4 & \textbf{0.5} & 0.6 & 0.7  \\
		\midrule\midrule
	
		OA(\%) & 90.75 & 90.80 & 90.83 & \textbf{90.85} & 90.82 & 90.79  \\
	
		mIoU(\%) & 82.37 & 82.51 & 82.53 & \textbf{82.56} & 82.52 & 82.49 \\
		
		\bottomrule 
	\end{tabular}	
\end{table}

\begin{table}[t]
	\centering	
	\caption{Comparisons of different weight parameters $\lambda_3$.}
	\label{Table 11}
		\begin{tabular}{ccccccc}
		\toprule
		\textbf{$\lambda_3$} & 0.2 & 0.3 & \textbf{0.4} & 0.5 & 0.6 & 0.7  \\
		\midrule\midrule
		
		OA(\%) & 90.72 & 90.76 & \textbf{90.79} & 90.75 & 90.70 & 90.68  \\
		
		mIoU(\%) & 82.33 & 82.38 & \textbf{82.48} & 82.36 & 82.31 & 82.27 \\
		
		\bottomrule 
	\end{tabular}	
\end{table}

\begin{table}[t]
	\centering	
	\caption{Ablation study for multiple loss functions.}
	\label{Table 12}
	\begin{tabular}{c|ccc|c|c}
		\toprule
		\textbf{Method} & \textbf{$\mathcal L_{ce}$} & \textbf{$\mathcal L_{cls}$} & \textbf{$\mathcal L_{aux}$} & \textbf{OA(\%)} & \textbf{mIoU(\%)} \\ 
		\midrule\midrule

		HMANet & \checkmark &  &  & 90.76 & 82.44 \\
		
		HMANet & \checkmark & \checkmark &  & 90.85 & 82.56 \\
		
		HMANet & \checkmark &  & \checkmark & 90.79 & 82.48 \\
		
		HMANet & \checkmark & \checkmark & \checkmark & \textbf{90.98} & \textbf{82.87} \\
		
		\bottomrule 
	\end{tabular}
\end{table}

\begin{table}[t]
	\centering	
	\caption{Ablation study for attention modules on Vaihingen test set. CAA represents class augmented attention module, CCA represents channel class attention module, RSA represents region shuffle attention module.}
	\label{Table 1}
	\scalebox{0.85}{
	\begin{tabular}{c|ccc|c|c}
		\toprule
		\textbf{Method} & \textbf{CAA} & \textbf{CCA} & \textbf{RSA} & \textbf{OA(\%)} & \textbf{mIoU(\%)} \\ 
		\midrule\midrule
		
		Baseline \cite{long2015fully} &  &  &  & 86.51 & 72.69 \\
		
		HMANet (VGG-16) & \checkmark &  &  & 89.15 & 79.56 \\
		
		HMANet (VGG-16) &  &  & \checkmark & 89.23 & 79.65 \\
		
		HMANet (VGG-16) & \checkmark & \checkmark &  & 89.58 & 80.24 \\
		
		HMANet (VGG-16) & \checkmark &  & \checkmark & 89.66 & 80.31 \\
		
		HMANet (VGG-16) & \checkmark & \checkmark & \checkmark & \textbf{89.95} & \textbf{80.68} \\
		
		\bottomrule 
	\end{tabular}}
\end{table}

\begin{table}[t]
	\centering	
	\caption{Comparison between different integration patterns. Cascade-C-R indicates that CAA embedded with CCA module is followed by RSA module, and vice versa. Parallel-C-R represents CAA embedded with CCA and RSA are appended on the top of the ResNet-101 in parallel.}
	\label{Table 2}
	\begin{tabular}{l|c|c}
		\toprule
		\textbf{Method} & \textbf{OA(\%)} & \textbf{mIoU(\%)} \\ 
		\midrule\midrule
		
		ResNet-101 Baseline &  90.12 & 80.81 \\
		
		ResNet-101 + Cascade-C-R & 90.88 & 82.62 \\
		
		ResNet-101 + Cascade-R-C &  90.76 & 82.45 \\
		
		ResNet-101 + Parallel-C-R  & \textbf{90.98} & \textbf{82.87} \\
		
		\bottomrule 
	\end{tabular}
\end{table}

\subsection{Experiments on Vaihingen Dataset}\label{section:4.4}

\subsubsection{Ablation Study for weight parameters and multiple loss functions}\label{section:4.4.7}

The proposed model utilizes multiple loss functions to optimize the learning process. We first conduct experiments to analyze the sensitivity of the model to the choice of the weight parameters $\lambda_2$ and $\lambda_3$. Concretely, we set the weight parameter of the main cross entropy loss ($i.e.$ $\lambda_1$) as 1 and only preserve one of the auxiliary loss functions to further study the optimal value of $\lambda_2$ and $\lambda_3$. The experimental results for $\lambda_2$ and $\lambda_3$ are presented in $\rm Tab.\ \ref{Table 10}$ and $\rm Tab.\ \ref{Table 11}$. It can be seen that the choice of $\lambda_2=0.5$ and $\lambda_3=0.4$ yield the best result, respectively. Besides, it is worth mentioning that the model is not particularly sensitive to parameter selection. Thus, in order to avoid the influence of the training error of each experiment, we conduct 5 experiments for each value of the parameters, and take the average value as the final result.

We further investigate the performance of the three loss functions following the optimal settings in $\rm Tab.\ \ref{Table 10}$ and $\rm Tab.\ \ref{Table 11}$. As shown in $\rm Tab.\ \ref{Table 12}$, both auxiliary loss functions have certain improvement effects on model training optimization. It yields a result of 90.98\% in overall accuracy and 82.87\% in mean IoU when we utilize all the loss functions.

\subsubsection{Ablation Study for Attention Modules}\label{section:4.4.1}
In the proposed HMANet, three attention modules are employed on the top of the dilation network to exploit global contextual representations from the perspective of space, channel and category. To further verify the performance of attention modules, we conduct extensive experiments with different settings in $\rm Tab.\ \ref{Table 1}$. Noted that for a fair comparison with the baseline model FCN (VGG-16), we also use VGG-16 as the backbone on HMANet in this experiment. Besides, we further investigate two integration patterns, that is, the parallel and cascading fashion, to adaptively accomplish information propagation.

As illustrated in $\rm Tab.\ \ref{Table 1}$, the proposed attention modules bring remarkable improvement compared with the baseline FCN (VGG-16). We can observe that the use of only class augmented attention module yields a result of 89.15\% in overall accuracy and 79.56\% in mean IoU, which brings 2.64\% and 6.87\% improvement in OA and mIoU, respectively. Meanwhile, employing region shuffle attention individually outperforms the baseline by 2.72\% in OA and 6.96\% in mIoU. Furthermore, when we employ the integration of two corresponding attention modules together, the performance of our network is further boosted up. Finally, it behaves superiorly compared to other methods when we integrate the three attention modules, which improves the segmentation performance over baseline by 3.44\% in OA and 7.99\% in mIoU. In summary, it can be seen that our approach brings great benefit to object segmentation via exploiting global context from different perspectives.

We further investigate the effect of different aggregation methods of the three attention modules. As shown in $\rm Tab.\ \ref{Table 2}$, the ResNet101 +Parallel-C-R, corresponding to the schematic diagram in $\rm Fig.\ \ref{Figure 2}$, achieve the best performance, $i.e.$, 90.98\% in overall accuracy, as well as 82.27\% in mean IoU. While the two cascading integration patterns, ``+Cascade-C-R'' and ``+Cascade-R-C'' achieve 90.88\% and 90.76\% in overall accuracy, respectively. It shows that the cascading integration patterns lead to a decline in experimental results. The reason may be that the region-wise attention representation is not conducive to the extraction of category information only in the case of direct serial connection.

\begin{table}
	\centering	
	\caption{Performance on Vaihingen test set for different ascending ratio $\alpha$ in CCA module.}
	\label{Table 3}
	\begin{tabular}{c|c|c}
		\toprule
		\textbf{Ratio} \bm{$\alpha$} & \textbf{OA(\%)} & \textbf{mIoU(\%)} \\ 
		\midrule\midrule
		
		50 & 90.78 & 82.48 \\
		
		75 & 90.80 & 82.49 \\
		
		100 &  90.82 & 82.52 \\
		
		125 & 90.84 & 82.53 \\
		
		\textbf{150} & \textbf{90.85} & \textbf{82.54} \\
		
		175 & 90.83 & 82.52 \\
		
		200 & 90.81 & 82.50 \\
		
		\bottomrule 
	\end{tabular}
\end{table}

\begin{table}[t]
	\centering	
	\caption{Effect of partition numbers $G_h$ and $G_w$ within region shuffle attention module.}
	\label{Table 4}
	\begin{tabular}{c|cc|c|c}
		\toprule
		\textbf{Method} & \bm{$G_h$} & \bm{$G_w$} & \textbf{OA(\%)} & \textbf{mIoU(\%)}\\ 
		\midrule\midrule
		ResNet-101 Baseline & - & - & 90.12 & 80.81 \\
		\midrule
		\multirow{7}{*}{RSA}
		
		& 16 & 16 & 90.70 & 82.35 \\
		
		& 16 & 8 & 90.75 & 82.44 \\
		
		& 8 & 16 & 90.77 & 82.47 \\
		
		& \textbf{8} & \textbf{8} & \textbf{90.79} & \textbf{82.49} \\
		
		& 8 & 4 & 90.78 & 82.47 \\
		
		& 4 & 8 & 86.76 & 82.46 \\
		
		& 4 & 4 & 90.75 & 82.44 \\
		
		\bottomrule 
	\end{tabular}
\end{table}

\subsubsection{Ablation Study for Sub-parameters}\label{section:4.4.2}

\noindent\textbf{Ascending ratio.} 
The ascending ratio $\alpha$ introduced in $\rm Eq.\ (\ref{Equation 4})$ is a hyper-parameter which allows us to control the scale of feature transformations. As the choice of ascending ratio does not have much effect on the computational cost, we only investigate the performance between a range of different $\alpha$ values. As shown in $\rm Tab.\ (\ref{Table 3})$, we can conclude that our approach consistently outperforms the baseline under different choices of hyper-parameters, among which the choice ratio $\alpha = 150$ achieves slightly better results than others. Qualitatively, the ratio $\alpha$ is the scaling factor of category information, which can takes a moderate value while controlling the computational cost. 

\begin{figure}[t]
	\centering
	\includegraphics[scale=0.45]{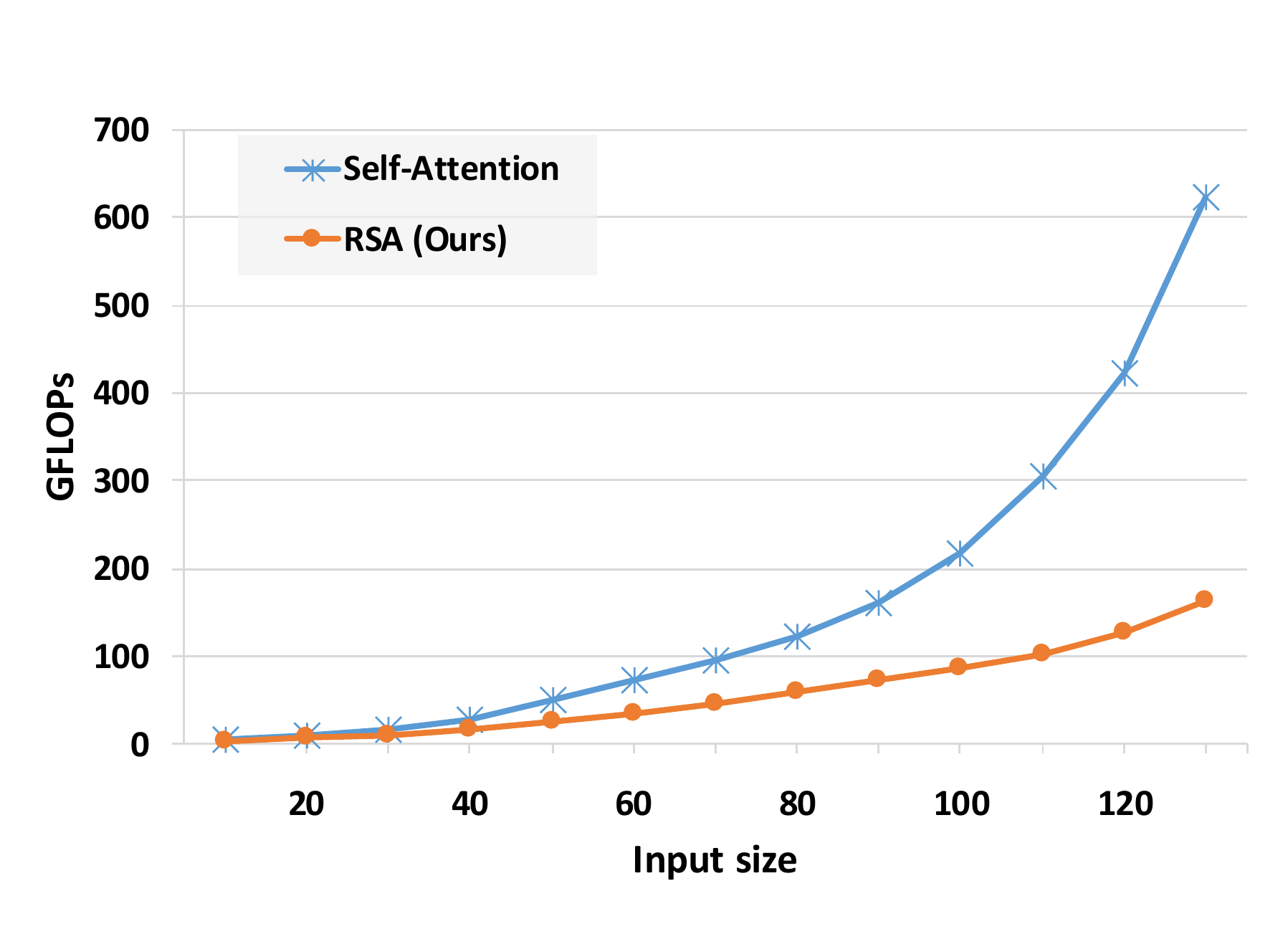}
	\caption{Comparison of numerical complexity. The $x$-axis represents the height and width of the input feature map and the $y$-axis represents the computation cost measured with GFLOPs.}
	\label{Figure 9}
\end{figure}

\noindent\textbf{Effect of the Partition numbers.} We further investigate the effect of different partition numbers of the proposed region shuffle attention module, $i.e.$, $G$ and $P$. We conduct extensive experiments with various choices of $G$ and $P$, and present the corresponding results in $\rm Tab.\ (\ref{Table 4})$. Noted that $G$ and $P$ are mutually constrained, namely, we just need to determine the values of $G_h$ and $G_w$. We can see that the performance is robust for a range of partition numbers, among which the choice $G_h=G_w=8$ achieve the best 90.79\% in overall accuracy and 82.49\% in mean IoU. Empirically, the output stride of the backbone is set to 8, that is, the height and width of the input feature is 64 pixels in our experiments, thus eclectic choice of grouping is more conducive to self-attention weighted representations of each region. In practice, using an identical partition number may not be optimal (due to the distinct roles performed by different base network and different training settings, $e.g.$, output stride and input size), so further improvements may be achievable by tuning the partition numbers to meet the needs of the given base architecture.

\begin{table}[t]
	\centering	
	\caption{Comparison with context aggregation approaches.}
	\label{Table 5}
	\begin{tabular}{l|c|c}
		\toprule
		\textbf{Method} & \textbf{OA(\%)} & \textbf{mIoU(\%)} \\ 
		\midrule\midrule
		
		ResNet-101 Baseline & 90.12 & 80.81 \\
		\midrule
		
		+ ASPP (Our impl.) \cite{chen2017rethinking} & 90.51 & 81.39 \\
		
		+ PPM (Our impl.) \cite{zhao2017pyramid} &  90.82 & 82.52 \\
		
		+ Self-Attention (Our impl.) \cite{wang2018non} & 90.62 & 82.17 \\
		
		+ RCCA (Our impl.) \cite{huang2019ccnet} & 90.76 & 82.45 \\
		
		\textbf{+ Ours} & \textbf{90.98} & \textbf{82.87} \\
		
		\bottomrule 
	\end{tabular}
\end{table}

\subsubsection{Comparison with Context Aggregation Approaches}\label{section:4.4.3}
We compare the performance of several well verified context aggregation approaches, $i.e.$, Atrous Spatial Pyramid Pooling (ASPP) in DeepLabv3 \cite{chen2017rethinking}, Pyramid Pooling Module (PPM) in PSPNet \cite{zhao2017pyramid}, RCCA in CCNet \cite{huang2019ccnet} and Self-Attention in non-local networks \cite{wang2018non}. All the experiments above are conducted under the same training/testing settings for fairness. We report the related results in $\rm Tab.\ (\ref{Table 5})$. Concretely, ``+PPM'' achieves better performance compared with ``+ASPP'' in terms of expanding local receptive fields. Both ``+Self-Attention'' and ``+RCCA'' generate contextual information from all spatial positions in the feature maps, leading to limited object contexts. In contrast, our HMANet calculates global correlations from the perspective of space, channel and category. Results show that HMANet outperforms other context aggregation approaches, which demonstrates the effectiveness of capturing global contextual information from different perspectives.

\subsubsection{Efficiency Comparison}\label{section:4.4.4}

\noindent\textbf{Comparison with Self-attention.} As illustrated in $\rm Fig.\ \ref{Figure 9}$. We first compare our RSA module with the standard self-attention mechanism in terms of the computation cost measured with GFLOPs. As the size of input feature map increases, the GFLOPs of self-attention mechanism gradually increases exponentially while the counterparts of our RSA module is almost linearly increasing. It can be seen that the RSA module is much more efficient than the self-attention mechanism when processing high-resolution feature maps.

\noindent\textbf{Comparison with Context Aggregation modules and Attention modules.} We further compare our proposed class augmented attention module and region shuffle attention module with ASPP \cite{chen2017deeplab,chen2017rethinking}, PPM \cite{zhao2017pyramid}, SA \cite{wang2018non}, RCCA \cite{huang2019ccnet}, OCR \cite{yuan2019object} and ISA \cite{huang2019interlaced} in terms of efficiency, including parameters, GPU memory and computation cost (GFLOPs). We report the results in $\rm Tab.\ (\ref{Table 6})$. Notably, we evaluate the cost of all above methods without considering the cost of backbone and include the cost of $3\times 3$ convolution for dimension reduction to ensure the fairness of the comparison. As shown in $\rm Tab.\ (\ref{Table 6})$, compared with standard Self-Attention (SA) mechanism, our RSA module requires $20\times$ less GPU memory usage and significantly reduce FLOPs by about 77\% with a few parameters, which proves the efficiency of region-wise representations in capturing long-range contextual information.

\begin{table}[t]
	\small
	\centering	
	\caption{Efficiency comparison with ASPP, PPM, Self-Attention, RCCA, OCR and ISA when processing input feature map of size [$1\times2048\times128\times128$] during inference stage.}
	\label{Table 6}
	\begin{tabular}{l|c|c|c}
		\toprule
		\textbf{Method} & \textbf{Params(M\large{$\blacktriangle$}}) & \textbf{Memory(MB\large{$\blacktriangle$}}) & \textbf{GFLOPs($\blacktriangle$}) \\ 
		\midrule\midrule
		
		ASPP \cite{chen2017rethinking} & 15.1 & 284 & 503 \\
		
		PPM \cite{zhao2017pyramid} & 22.0 & 792 & 619 \\
		
		SA \cite{wang2018non} & 10.5 & 2168 & 619 \\
		
		RCCA \cite{huang2019ccnet} & 10.6 & 427 & 804 \\
		
		OCR \cite{yuan2019object} & 10.5 & 202 & 354 \\
		
		ISA \cite{huang2019interlaced} & 11.8 & 252 & 386 \\
		
		CAA(Ours) & 9.3 & 283 & 148 \\
		
		\textbf{RSA(Ours)} & \textbf{3.8} & \textbf{110} & \textbf{144} \\
		
		\bottomrule 
	\end{tabular}
\end{table}

\begin{table}[t]
	\centering	
	\caption{Performace comparison between data augmentation (DA), multi-grid (MG) and multi-scale with horizontal flipping (MS + Flip). We report the results on the test set of Vaihingen.}
	\label{Table 7}
	\begin{tabular}{c|ccc|c|c}
		\toprule
		\textbf{Method} & \textbf{DA} & \textbf{MG} & \textbf{MS + Flip} & \textbf{OA(\%)} & \textbf{mIoU(\%)} \\ 
		\midrule\midrule
		HMANet &  &  &  & 90.98 & 82.87 \\
		
		HMANet & \checkmark &  &  & 91.17 & 83.11 \\
		
		HMANet & \checkmark & \checkmark &  & 91.28 & 83.27 \\
		
		HMANet & \checkmark & \checkmark & \checkmark & \textbf{91.44} & \textbf{83.49} \\
		
		\bottomrule 
	\end{tabular}
\end{table}

\subsubsection{Comparison with State-of-the-art}\label{section:4.4.5}

We first adopt some common strategies to improve performance following \cite{fu2019dual, yuan2018ocnet, li2019expectation}. (1) DA: Data augmentation with random scaling (from 0.5 to 2.0) and random left-right flipping. (2) Multi-Grid: We employ hierarchical grids of different sizes (1,2,4) within stage-4 of ResNet-101. (3) MS + Flip: We average the segmentation score maps from 5 image scales $\left\{ 0.5, 0.75, 1.0, 1.25, 1.5 \right\} $ and left-right flipping counterparts during inference.

Experimental results are shown in $\rm Tab.\ (\ref{Table 7})$. We successively adopt the above strategies to obtain better object representations, which achieves 0.19\% ,0.11\% and 0.16\% improvements respectively in overall accuracy.

We further compare our method with existing methods on Vaihingen test set. Notably, most of the methods adopt ResNet-101 as their backbone. Results are shown in $\rm Tab.\ (\ref{Table 8})$. It can be seen that our HMANet (ResNet-101) outperforms other context aggregation methods and attention-based methods by a large margin. Moreover, our HMANet is much more efficient in parameters, memory and GFLOPs. Especially, our $F_1$ score of Car is much higher than other approaches, it improves the second best CCNet by 0.93\%, which demonstrates the effectiveness of capturing category-based information and global region-wise correlation.

\subsubsection{Visualization Results}\label{section:4.4.6}
\begin{figure*}
	\centering
	\includegraphics[scale=0.191]{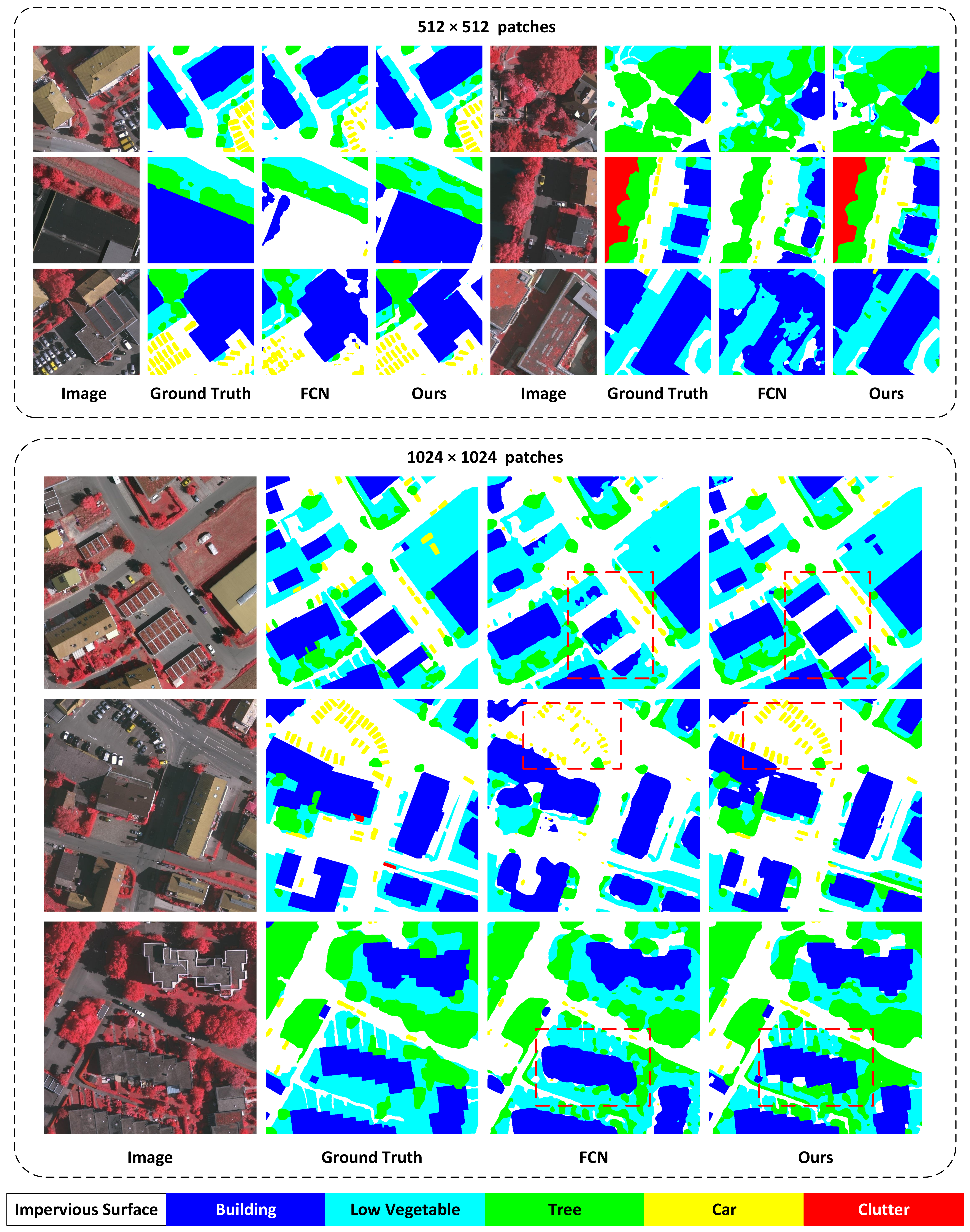}
	\caption{Qualitative comparisons between our method and baseline on Vaihingen test set.}
	\label{Figure 6}
\end{figure*}

We provide qualitative comparisons between our HMANet and baseline network in $\rm Fig.\ \ref{Figure 6}$, including $512\times512$ and $1024\times1024$ patches. In particular, we leverage the red dashed box to mark those challenging regions that are easily to be misclassified. It can be seen that our method outperforms the baseline by a large margin. HMANet predicts more accurate segmentation maps, that is, it can obtain finer boundary information and maintain the object coherence, which demonstrates the effectiveness of modeling category-based correlation and region-wise representations.

\begin{table*}[t]
	\centering	
	\caption{Comparisons with state-of-the-arts on Vaihingen test set.}
	\label{Table 8}
	\begin{tabular}{l|c|ccccc|c|c|c}
		\toprule
		\textbf{Method} & \textbf{Backbone} & \textbf{Imp. surf.} & \textbf{Building} & \textbf{Low veg.} & \textbf{Tree} & \textbf{Car} & \textbf{mean $\bm{F_1}$} & \textbf{OA(\%)} & \textbf{mIoU(\%)} \\ 
		\midrule\midrule
		
		FCN \cite{long2015fully} & VGG-16 & 88.67 & 92.83 & 76.32 & 86.67 & 74.21 & 83.74 & 86.51 & 72.69  \\
		
		UZ\_1 \cite{volpi2016dense} & - & 89.20 & 92.50 & 81.60 & 86.90 & 57.30 & 81.50 & 87.30 & - \\
		
		RoteEqNet \cite{marcos2018land} & - & 89.50 & 94.80 & 77.50 & 86.50 & 72.60 & 84.18 & 87.50 & -  \\
		
		S-RA-FCN \cite{mou2019relation} & VGG-16 & 91.47 & 94.97 & 80.63 & 88.57 & 87.05 & 88.54 & 89.23 & 79.76  \\
		
		UFMG\_4 \cite{nogueira2019dynamic} & - & 91.10 & 94.50 & 82.90 & 88.80 & 81.30 & 87.72 & 89.40 & -  \\
		
		V-FuseNet \cite{audebert2018beyond} & - & 92.00 & 94.40 & 84.50 & 89.90 & 86.30 & 89.42 & 90.00 & - \\
		
		DLR\_9 \cite{marmanis2018classification} & - & 92.40 & 95.20 & 83.90 & 89.90 & 81.20 & 88.52 & 90.30 & - \\
		
		TreeUNet \cite{yue2019treeunet} & - & 92.50 & 94.90 & 83.60 & 89.60 & 85.90 & 89.30 & 90.40 & - \\
		
		DANet \cite{fu2019dual} & ResNet-101 & 91.63 & 95.02 & 83.25 & 88.87 & 87.16 & 89.19 & 90.44 & 81.32  \\
		
		DeepLabV3+ \cite{chen2017rethinking} & ResNet-101 & 92.38 & 95.17 & 84.29 & 89.52 & 86.47 & 89.57 & 90.56 & 81.47  \\
		
		PSPNet \cite{zhao2017pyramid} & ResNet-101 & 92.79 & 95.46 & 84.51 & 89.94 & 88.61 & 90.26 & 90.85 & 82.58  \\
		
		ACFNet \cite{zhang2019acfnet} & ResNet-101 & 92.93 & 95.27 & 84.46 & 90.05 & 88.64 & 90.27 & 90.90 & 82.68  \\
		
		BKHN11 & ResNet-101 & 92.90 & \textbf{96.00} & 84.60 & 89.90 & 88.60 & 90.40 & 91.00 & - \\
		
		CASIA2 \cite{liu2018semantic} & ResNet-101 & 93.20 & \textbf{96.00} & 84.70 & 89.90 & 86.70 & 90.10 & 91.10 & - \\
		
		CCNet \cite{huang2019ccnet} & ResNet-101 & 93.29 & 95.53 & 85.06 & 90.34 & 88.70 & 90.58 & 91.11 & 82.76  \\
		
		\midrule
		
		\textbf{HMANet (Ours)} & VGG-16 & 91.86 & 94.52 & 83.17 & 89.81 & 87.15 & 89.30 & 89.95 & 80.68  \\
		
		\textbf{HMANet (Ours)} & ResNet-101 & \textbf{93.50} & 95.86 & \textbf{85.41} & \textbf{90.40} & \textbf{89.63} & \textbf{90.96} & \textbf{91.44} & \textbf{83.49}  \\
		
		\bottomrule 
	\end{tabular}
\end{table*}

\subsection{Experiments on Potsdam Dataset}\label{section:4.5}

\begin{table*}[!h]
	\centering	
	\caption{Numerical comparisons with state-of-the-arts on Potsdam test set.}
	\label{Table 9}
	\begin{tabular}{l|c|ccccc|c|c|c}
		\toprule
		\textbf{Method} & \textbf{Backbone} & \textbf{Imp. surf.} & \textbf{Building} & \textbf{Low veg.} & \textbf{Tree} & \textbf{Car} & \textbf{mean $\bm{F_1}$} & \textbf{OA(\%)} & \textbf{mIoU(\%)} \\ 
		\midrule\midrule
		
		FCN \cite{long2015fully} & VGG-16 & 88.61 & 93.29 & 83.29 & 79.83 & 93.02 & 87.61 & 85.59 & 78.34  \\
		
		UZ\_1 \cite{volpi2016dense} & - & 89.30 & 95.40 & 81.80 & 80.50 & 86.50 & 86.70 & 85.80 & - \\
		
		UFMG\_4 \cite{nogueira2019dynamic} & - & 90.80 & 95.60 & 84.40 & 84.30 & 92.40 & 89.50 & 87.90 & -  \\
		
		S-RA-FCN \cite{mou2019relation} & VGG-16 & 91.33 & 94.70 & 86.81 & 83.47 & 94.52 & 90.17 & 88.59 & 82.38  \\
		
		V-FuseNet \cite{audebert2018beyond} & - & 92.70 & 96.30 & 87.30 & 88.50 & 95.40 & 92.04 & 90.60 & - \\
		
		TSMTA \cite{ding2020semantic} & ResNet-101 & 92.91 & 97.13 & 87.03 & 87.26 & 95.16 & 91.90 & 90.64 & - \\
		
		Multi-filter CNN \cite{sun2018developing} & VGG-16 & 90.94 & 96.98 & 76.32 & 73.37 & 88.55 & 85.23 & 90.65 & - \\
		
		TreeUNet \cite{yue2019treeunet} & - & 93.10 & 97.30 & 86.60 & 87.10 & 95.80 & 91.98 & 90.70 & - \\
		
		DeepLabV3+ \cite{chen2017rethinking} & ResNet-101 & 92.95 & 95.88 & 87.62 & 88.15 & 96.02 & 92.12 & 90.88 & 84.32  \\
		
		CASIA3 \cite{liu2018semantic} & ResNet-101 & 93.40 & 96.80 & 87.60 & 88.30 & 96.10 & 92.44 & 91.00 & - \\
		
		PSPNet \cite{zhao2017pyramid} & ResNet-101 & 93.36 & 96.97 & 87.75 & 88.50 & 95.42 & 92.40 & 91.08 & 84.88  \\
		
		BKHN3 & ResNet-101 & 93.30 & 97.20 & 88.00 & 88.50 & 96.00 & 92.60 & 91.10 & - \\
		
		AMA\_1 & - & 93.40 & 96.80 & 87.70 & 88.80 & 96.00 & 92.54 & 91.20 & - \\
		
		CCNet \cite{huang2019ccnet} & ResNet-101 & 93.58 & 96.77 & 86.87 & 88.59 & 96.24 & 92.41 & 91.47 & 85.65  \\
		
		HUSTW4 \cite{sun2019problems} & - & 93.60 & \textbf{97.60} & 88.50 & 88.80 & 94.60 & 92.62 & 91.60 & - \\
		
		SWJ\_2 & ResNet-101 & \textbf{94.40} & 97.40 & 87.80 & 87.60 & 94.70 & 92.38 & 91.70 & - \\
		
		\midrule
		
		\textbf{HMANet (Ours)} & VGG-16 & 92.38 & 96.08 & 86.93 & 88.21 & 95.44 & 91.81 & 90.46 & 83.53  \\
		
		\textbf{HMANet (Ours)} & ResNet-101 & 93.85 & 97.56 & \textbf{88.65} & \textbf{89.12} & \textbf{96.84} & \textbf{93.20} & \textbf{92.21} & \textbf{87.28}  \\
		
		\bottomrule 
	\end{tabular}
\end{table*}

\begin{figure*}
	\centering
	\includegraphics[scale=0.2]{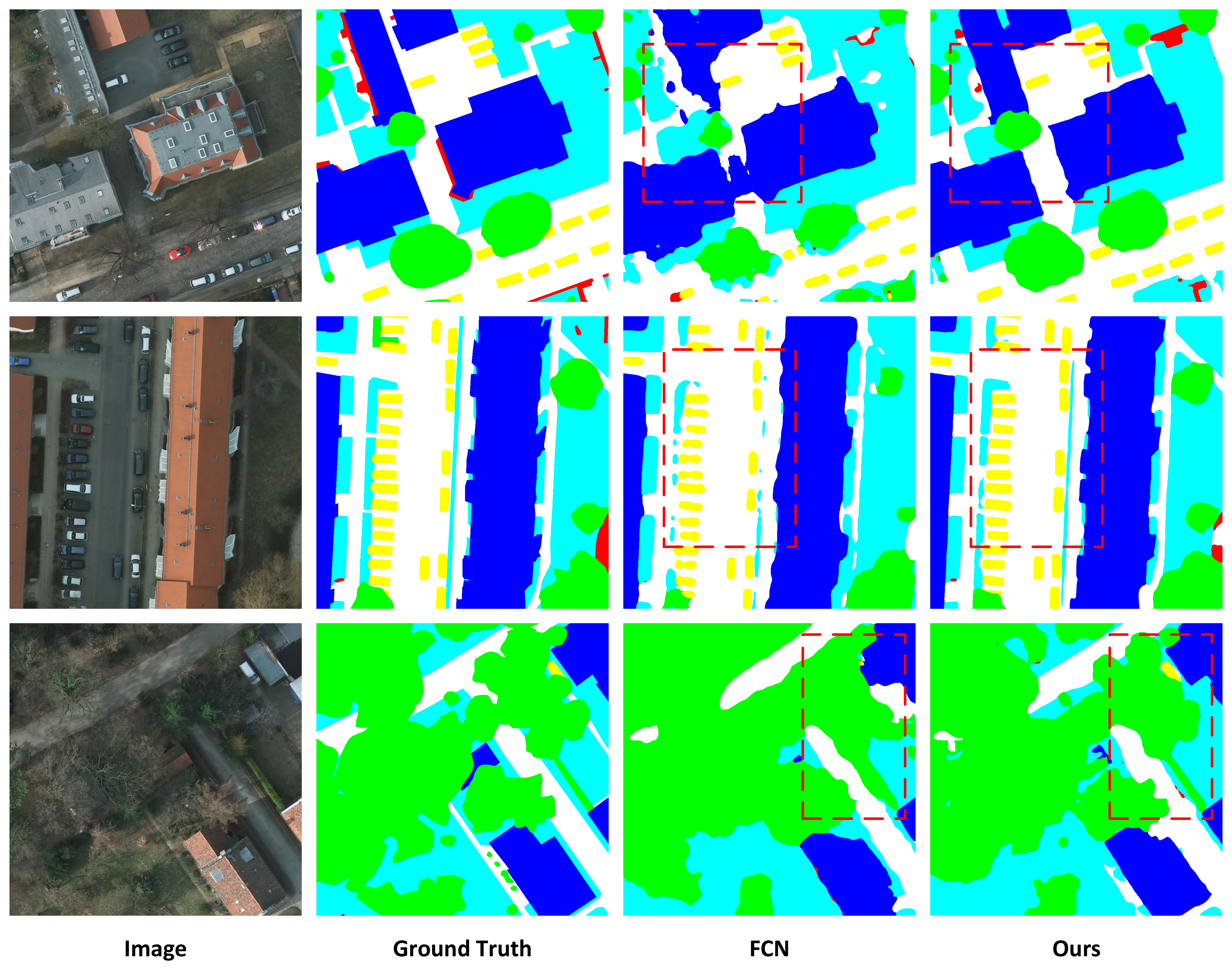}
	\caption{Visualization results of HMANet on Potsdam test set.}
	\label{Figure 7}
\end{figure*}

We carry out experiments on ISPRS Potsdam benchmark to further evaluate the effectiveness of HMANet. Empirically, we adopt the same training and testing settings on Potsdam dataset. Numerical comparisons with state-of-the-art methods are shown in $\rm Tab.\ (\ref{Table 9})$. Remarkably, HMANet (ResNet-101) achieve 92.21\% in overall accuracy and 87.28\% in mean IoU. Notably, we compare the two types of available input images, $i.e.$, RGB and IRRG color modes. Results show that the former can obtain better segmentation maps.

In addition, qualitative results are presented in $\rm Fig.\ \ref{Figure 7}$. It can be seen that HMANet produces better segmentation maps than baseline. We mark the improved regions with red dashed boxes (Best viewed in color).

\section{Conclusion}\label{section:5}In this paper, we propose a novel attention-based framework for dense prediction tasks in the field of remote sensing, namely Hybrid Multiple Attention Network (HMANet), which adaptively captures global contextual information from the perspective of space, channel and category. In particular, we introduce a class augmented attention module embedded with a class channel attention module to compute category-based correlation and further adaptively recalibrate the class-level information. Additionally, to address the feature redundancy and improve the efficiency of self-attention mechanism, a region shuffle attention module is presented to obtain robust region-wise representations. Extensive experiments on ISPRS Vaihingen and Potsdam benchmark demonstrate the effectiveness and efficiency of the proposed HMANet.

\bibliographystyle{IEEEtran} 
\bibliography{main}

\begin{thebibliography}{10}
\providecommand{\url}[1]{#1}
\csname url@samestyle\endcsname
\providecommand{\newblock}{\relax}
\providecommand{\bibinfo}[2]{#2}
\providecommand{\BIBentrySTDinterwordspacing}{\spaceskip=0pt\relax}
\providecommand{\BIBentryALTinterwordstretchfactor}{4}
\providecommand{\BIBentryALTinterwordspacing}{\spaceskip=\fontdimen2\font plus
\BIBentryALTinterwordstretchfactor\fontdimen3\font minus
  \fontdimen4\font\relax}
\providecommand{\BIBforeignlanguage}[2]{{%
\expandafter\ifx\csname l@#1\endcsname\relax
\typeout{** WARNING: IEEEtran.bst: No hyphenation pattern has been}%
\typeout{** loaded for the language `#1'. Using the pattern for}%
\typeout{** the default language instead.}%
\else
\language=\csname l@#1\endcsname
\fi
#2}}
\providecommand{\BIBdecl}{\relax}
\BIBdecl

\bibitem{maboudi2018integrating}
M.~Maboudi, J.~Amini, S.~Malihi, and M.~Hahn, ``Integrating fuzzy object based
  image analysis and ant colony optimization for road extraction from remotely
  sensed images,'' \emph{ISPRS Journal of Photogrammetry and Remote Sensing},
  vol. 138, pp. 151--163, 2018.

\bibitem{zhang2011mapping}
Q.~Zhang and K.~C. Seto, ``Mapping urbanization dynamics at regional and global
  scales using multi-temporal dmsp/ols nighttime light data,'' \emph{Remote
  Sensing of Environment}, vol. 115, no.~9, pp. 2320--2329, 2011.

\bibitem{marcos2018land}
D.~Marcos, M.~Volpi, B.~Kellenberger, and D.~Tuia, ``Land cover mapping at very
  high resolution with rotation equivariant cnns: Towards small yet accurate
  models,'' \emph{ISPRS journal of photogrammetry and remote sensing}, vol.
  145, pp. 96--107, 2018.

\bibitem{pal2005random}
M.~Pal, ``Random forest classifier for remote sensing classification,''
  \emph{International Journal of Remote Sensing}, vol.~26, no.~1, pp. 217--222,
  2005.

\bibitem{gualtieri1999support}
J.~A. Gualtieri and R.~F. Cromp, ``Support vector machines for hyperspectral
  remote sensing classification,'' in \emph{27th AIPR Workshop: Advances in
  Computer-Assisted Recognition}, vol. 3584.\hskip 1em plus 0.5em minus
  0.4em\relax International Society for Optics and Photonics, 1999, pp.
  221--232.

\bibitem{zhong2007multiple}
P.~Zhong and R.~Wang, ``A multiple conditional random fields ensemble model for
  urban area detection in remote sensing optical images,'' \emph{IEEE
  Transactions on Geoscience and Remote Sensing}, vol.~45, no.~12, pp.
  3978--3988, 2007.

\bibitem{long2015fully}
J.~Long, E.~Shelhamer, and T.~Darrell, ``Fully convolutional networks for
  semantic segmentation,'' in \emph{Proceedings of the IEEE conference on
  computer vision and pattern recognition}, 2015, pp. 3431--3440.

\bibitem{chen2017rethinking}
L.-C. Chen, G.~Papandreou, F.~Schroff, and H.~Adam, ``Rethinking atrous
  convolution for semantic image segmentation,'' \emph{arXiv preprint
  arXiv:1706.05587}, 2017.

\bibitem{zhao2017pyramid}
H.~Zhao, J.~Shi, X.~Qi, X.~Wang, and J.~Jia, ``Pyramid scene parsing network,''
  in \emph{Proceedings of the IEEE conference on computer vision and pattern
  recognition}, 2017, pp. 2881--2890.

\bibitem{liu2018semantic}
Y.~Liu, B.~Fan, L.~Wang, J.~Bai, S.~Xiang, and C.~Pan, ``Semantic labeling in
  very high resolution images via a self-cascaded convolutional neural
  network,'' \emph{ISPRS Journal of Photogrammetry and Remote Sensing}, vol.
  145, pp. 78--95, 2018.

\bibitem{wang2018non}
X.~Wang, R.~Girshick, A.~Gupta, and K.~He, ``Non-local neural networks,'' in
  \emph{Proceedings of the IEEE Conference on Computer Vision and Pattern
  Recognition}, 2018, pp. 7794--7803.

\bibitem{fu2019dual}
J.~Fu, J.~Liu, H.~Tian, Y.~Li, Y.~Bao, Z.~Fang, and H.~Lu, ``Dual attention
  network for scene segmentation,'' in \emph{Proceedings of the IEEE Conference
  on Computer Vision and Pattern Recognition}, 2019, pp. 3146--3154.

\bibitem{huang2019ccnet}
Z.~Huang, X.~Wang, L.~Huang, C.~Huang, Y.~Wei, and W.~Liu, ``Ccnet: Criss-cross
  attention for semantic segmentation,'' in \emph{Proceedings of the IEEE
  International Conference on Computer Vision}, 2019, pp. 603--612.

\bibitem{li2019expectation}
X.~Li, Z.~Zhong, J.~Wu, Y.~Yang, Z.~Lin, and H.~Liu, ``Expectation-maximization
  attention networks for semantic segmentation,'' in \emph{Proceedings of the
  IEEE International Conference on Computer Vision}, 2019, pp. 9167--9176.

\bibitem{huang2019interlaced}
L.~Huang, Y.~Yuan, J.~Guo, C.~Zhang, X.~Chen, and J.~Wang, ``Interlaced sparse
  self-attention for semantic segmentation,'' \emph{arXiv preprint
  arXiv:1907.12273}, 2019.

\bibitem{chen20182}
Y.~Chen, Y.~Kalantidis, J.~Li, S.~Yan, and J.~Feng, ``A\^{} 2-nets: Double
  attention networks,'' in \emph{Advances in Neural Information Processing
  Systems}, 2018, pp. 352--361.

\bibitem{chen2017deeplab}
L.-C. Chen, G.~Papandreou, I.~Kokkinos, K.~Murphy, and A.~L. Yuille, ``Deeplab:
  Semantic image segmentation with deep convolutional nets, atrous convolution,
  and fully connected crfs,'' \emph{IEEE transactions on pattern analysis and
  machine intelligence}, vol.~40, no.~4, pp. 834--848, 2017.

\bibitem{mi2020superpixel}
L.~Mi and Z.~Chen, ``Superpixel-enhanced deep neural forest for remote sensing
  image semantic segmentation,'' \emph{ISPRS Journal of Photogrammetry and
  Remote Sensing}, vol. 159, pp. 140--152, 2020.

\bibitem{he2016deep}
K.~He, X.~Zhang, S.~Ren, and J.~Sun, ``Deep residual learning for image
  recognition,'' in \emph{Proceedings of the IEEE conference on computer vision
  and pattern recognition}, 2016, pp. 770--778.

\bibitem{huang2017densely}
G.~Huang, Z.~Liu, L.~Van Der~Maaten, and K.~Q. Weinberger, ``Densely connected
  convolutional networks,'' in \emph{Proceedings of the IEEE conference on
  computer vision and pattern recognition}, 2017, pp. 4700--4708.

\bibitem{russakovsky2015imagenet}
O.~Russakovsky, J.~Deng, H.~Su, J.~Krause, S.~Satheesh, S.~Ma, Z.~Huang,
  A.~Karpathy, A.~Khosla, M.~Bernstein \emph{et~al.}, ``Imagenet large scale
  visual recognition challenge,'' \emph{International journal of computer
  vision}, vol. 115, no.~3, pp. 211--252, 2015.

\bibitem{ronneberger2015u}
O.~Ronneberger, P.~Fischer, and T.~Brox, ``U-net: Convolutional networks for
  biomedical image segmentation,'' in \emph{International Conference on Medical
  image computing and computer-assisted intervention}.\hskip 1em plus 0.5em
  minus 0.4em\relax Springer, 2015, pp. 234--241.

\bibitem{lin2017refinenet}
G.~Lin, A.~Milan, C.~Shen, and I.~Reid, ``Refinenet: Multi-path refinement
  networks for high-resolution semantic segmentation,'' in \emph{Proceedings of
  the IEEE conference on computer vision and pattern recognition}, 2017, pp.
  1925--1934.

\bibitem{yu2018learning}
C.~Yu, J.~Wang, C.~Peng, C.~Gao, G.~Yu, and N.~Sang, ``Learning a
  discriminative feature network for semantic segmentation,'' in
  \emph{Proceedings of the IEEE Conference on Computer Vision and Pattern
  Recognition}, 2018, pp. 1857--1866.

\bibitem{badrinarayanan2017segnet}
V.~Badrinarayanan, A.~Kendall, and R.~Cipolla, ``Segnet: A deep convolutional
  encoder-decoder architecture for image segmentation,'' \emph{IEEE
  transactions on pattern analysis and machine intelligence}, vol.~39, no.~12,
  pp. 2481--2495, 2017.

\bibitem{chen2018encoder}
L.-C. Chen, Y.~Zhu, G.~Papandreou, F.~Schroff, and H.~Adam, ``Encoder-decoder
  with atrous separable convolution for semantic image segmentation,'' in
  \emph{Proceedings of the European conference on computer vision (ECCV)},
  2018, pp. 801--818.

\bibitem{cheng2019spgnet}
B.~Cheng, L.-C. Chen, Y.~Wei, Y.~Zhu, Z.~Huang, J.~Xiong, T.~S. Huang, W.-M.
  Hwu, and H.~Shi, ``Spgnet: Semantic prediction guidance for scene parsing,''
  in \emph{Proceedings of the IEEE International Conference on Computer
  Vision}, 2019, pp. 5218--5228.

\bibitem{peng2017large}
C.~Peng, X.~Zhang, G.~Yu, G.~Luo, and J.~Sun, ``Large kernel matters--improve
  semantic segmentation by global convolutional network,'' in \emph{Proceedings
  of the IEEE conference on computer vision and pattern recognition}, 2017, pp.
  4353--4361.

\bibitem{yu2018bisenet}
C.~Yu, J.~Wang, C.~Peng, C.~Gao, G.~Yu, and N.~Sang, ``Bisenet: Bilateral
  segmentation network for real-time semantic segmentation,'' in
  \emph{Proceedings of the European Conference on Computer Vision (ECCV)},
  2018, pp. 325--341.

\bibitem{mou2019relation}
L.~Mou, Y.~Hua, and X.~X. Zhu, ``Relation matters: Relational context-aware
  fully convolutional network for semantic segmentation of high-resolution
  aerial images,'' \emph{IEEE Transactions on Geoscience and Remote Sensing},
  2020.

\bibitem{yue2019treeunet}
K.~Yue, L.~Yang, R.~Li, W.~Hu, F.~Zhang, and W.~Li, ``Treeunet: Adaptive tree
  convolutional neural networks for subdecimeter aerial image segmentation,''
  \emph{ISPRS Journal of Photogrammetry and Remote Sensing}, vol. 156, pp.
  1--13, 2019.

\bibitem{marmanis2018classification}
D.~Marmanis, K.~Schindler, J.~D. Wegner, S.~Galliani, M.~Datcu, and U.~Stilla,
  ``Classification with an edge: Improving semantic image segmentation with
  boundary detection,'' \emph{ISPRS Journal of Photogrammetry and Remote
  Sensing}, vol. 135, pp. 158--172, 2018.

\bibitem{cao2019end}
Z.~Cao, K.~Fu, X.~Lu, W.~Diao, H.~Sun, M.~Yan, H.~Yu, and X.~Sun, ``End-to-end
  dsm fusion networks for semantic segmentation in high-resolution aerial
  images,'' \emph{IEEE Geoscience and Remote Sensing Letters}, 2019.

\bibitem{bahdanau2014neural}
D.~Bahdanau, K.~Cho, and Y.~Bengio, ``Neural machine translation by jointly
  learning to align and translate,'' \emph{arXiv preprint arXiv:1409.0473},
  2014.

\bibitem{vaswani2017attention}
A.~Vaswani, N.~Shazeer, N.~Parmar, J.~Uszkoreit, L.~Jones, A.~N. Gomez,
  {\L}.~Kaiser, and I.~Polosukhin, ``Attention is all you need,'' in
  \emph{Advances in neural information processing systems}, 2017, pp.
  5998--6008.

\bibitem{hu2018squeeze}
J.~Hu, L.~Shen, and G.~Sun, ``Squeeze-and-excitation networks,'' in
  \emph{Proceedings of the IEEE conference on computer vision and pattern
  recognition}, 2018, pp. 7132--7141.

\bibitem{zhang2019acfnet}
F.~Zhang, Y.~Chen, Z.~Li, Z.~Hong, J.~Liu, F.~Ma, J.~Han, and E.~Ding,
  ``Acfnet: Attentional class feature network for semantic segmentation,'' in
  \emph{Proceedings of the IEEE International Conference on Computer Vision},
  2019, pp. 6798--6807.

\bibitem{yuan2019object}
Y.~Yuan, X.~Chen, and J.~Wang, ``Object-contextual representations for semantic
  segmentation,'' \emph{arXiv preprint arXiv:1909.11065}, 2019.

\bibitem{zhu2019asymmetric}
Z.~Zhu, M.~Xu, S.~Bai, T.~Huang, and X.~Bai, ``Asymmetric non-local neural
  networks for semantic segmentation,'' in \emph{Proceedings of the IEEE
  International Conference on Computer Vision}, 2019, pp. 593--602.

\bibitem{wang2019eca}
Q.~Wang, B.~Wu, P.~Zhu, P.~Li, W.~Zuo, and Q.~Hu, ``Eca-net: Efficient channel
  attention for deep convolutional neural networks,'' \emph{arXiv preprint
  arXiv:1910.03151}, 2019.

\bibitem{vaihingen}
\BIBentryALTinterwordspacing
Isprs.2d semantic labeling contest-vaihingen. [Online]. Available:
  \url{http://www2.isprs.org/commissions/comm3/wg4/2d-sem-label-vaihingen.html}
\BIBentrySTDinterwordspacing

\bibitem{potsdam}
\BIBentryALTinterwordspacing
Isprs.2d semantic labeling contest-potsdam. [Online]. Available:
  \url{http://www2.isprs.org/commissions/comm3/wg4/2d-sem-label-potsdam.html}
\BIBentrySTDinterwordspacing

\bibitem{maggiori2017high}
E.~Maggiori, Y.~Tarabalka, G.~Charpiat, and P.~Alliez, ``High-resolution aerial
  image labeling with convolutional neural networks,'' \emph{IEEE Transactions
  on Geoscience and Remote Sensing}, vol.~55, no.~12, pp. 7092--7103, 2017.

\bibitem{volpi2016dense}
M.~Volpi and D.~Tuia, ``Dense semantic labeling of subdecimeter resolution
  images with convolutional neural networks,'' \emph{IEEE Transactions on
  Geoscience and Remote Sensing}, vol.~55, no.~2, pp. 881--893, 2016.

\bibitem{sherrah2016fully}
J.~Sherrah, ``Fully convolutional networks for dense semantic labelling of
  high-resolution aerial imagery,'' \emph{arXiv preprint arXiv:1606.02585},
  2016.

\bibitem{rota2018place}
S.~Rota~Bul{\`o}, L.~Porzi, and P.~Kontschieder, ``In-place activated batchnorm
  for memory-optimized training of dnns,'' in \emph{Proceedings of the IEEE
  Conference on Computer Vision and Pattern Recognition}, 2018, pp. 5639--5647.

\bibitem{yuan2018ocnet}
Y.~Yuan and J.~Wang, ``Ocnet: Object context network for scene parsing,''
  \emph{arXiv preprint arXiv:1809.00916}, 2018.

\bibitem{nogueira2019dynamic}
K.~Nogueira, M.~Dalla~Mura, J.~Chanussot, W.~R. Schwartz, and J.~A. dos Santos,
  ``Dynamic multicontext segmentation of remote sensing images based on
  convolutional networks,'' \emph{IEEE Transactions on Geoscience and Remote
  Sensing}, vol.~57, no.~10, pp. 7503--7520, 2019.

\bibitem{audebert2018beyond}
N.~Audebert, B.~Le~Saux, and S.~Lef{\`e}vre, ``Beyond rgb: Very high resolution
  urban remote sensing with multimodal deep networks,'' \emph{ISPRS Journal of
  Photogrammetry and Remote Sensing}, vol. 140, pp. 20--32, 2018.

\bibitem{ding2020semantic}
L.~Ding, J.~Zhang, and L.~Bruzzone, ``Semantic segmentation of large-size vhr
  remote sensing images using a two-stage multiscale training architecture,''
  \emph{IEEE Transactions on Geoscience and Remote Sensing}, 2020.

\bibitem{sun2018developing}
Y.~Sun, X.~Zhang, Q.~Xin, and J.~Huang, ``Developing a multi-filter
  convolutional neural network for semantic segmentation using high-resolution
  aerial imagery and lidar data,'' \emph{ISPRS journal of photogrammetry and
  remote sensing}, vol. 143, pp. 3--14, 2018.

\bibitem{sun2019problems}
Y.~Sun, Y.~Tian, and Y.~Xu, ``Problems of encoder-decoder frameworks for
  high-resolution remote sensing image segmentation: Structural stereotype and
  insufficient learning,'' \emph{Neurocomputing}, vol. 330, pp. 297--304, 2019.

\end{thebibliography}

\end{document}